\documentclass[journal]{IEEEtran}
\usepackage{amsmath,amsfonts}
\usepackage{algorithmic}
\usepackage{algorithm}
\usepackage{array}
\usepackage[caption=false,font=normalsize,labelfont=sf,textfont=sf]{subfig}
\usepackage{textcomp}
\usepackage{stfloats}
\usepackage{url}
\usepackage{verbatim}
\usepackage{graphicx}
\usepackage{cite}
\usepackage{times}
\usepackage{epsfig}
\usepackage{graphicx}
\usepackage{amssymb}
\usepackage{fontawesome}
\usepackage{multirow}
\usepackage{tabularx}
\begin{document}

\title{Prompt-based Ingredient-Oriented All-in-One Image Restoration}

\author{Hu Gao and Depeng Dang
\thanks{Hu Gao and Depeng Dang are with the School of
Artificial Intelligence, Beijing Normal University,
Beijing 100000, China (e-mail: gao\_h@mail.bnu.edu.cn, ddepeng@bnu.edu.cn).}}



\maketitle

\begin{abstract}
Image restoration aims to recover the high-quality images from their degraded observations. Since most existing methods have been dedicated into single degradation removal, they may not yield optimal results on other types of degradations, which do not satisfy the applications in real world scenarios. In this paper, we propose a novel data ingredient-oriented approach that leverages prompt-based learning to enable a single model to efficiently tackle multiple image degradation tasks. Specifically, we utilize a encoder to capture features and introduce  prompts with degradation-specific information to guide the decoder in adaptively recovering images affected by various degradations. In order to model the local invariant properties and non-local information for high-quality image restoration, we combined CNNs operations
and Transformers. Simultaneously, we made several key designs in the Transformer blocks (multi-head rearranged attention with prompts and simple-gate feed-forward network) to reduce computational requirements and selectively determines what information should be persevered to facilitate efficient recovery of potentially sharp images. Furthermore, we incorporate a feature fusion mechanism further explores the multi-scale information  to improve the aggregated features. The resulting tightly interlinked hierarchy architecture, named as CAPTNet, extensive experiments demonstrate that our method performs competitively to the  state-of-the-art.

\end{abstract}

\begin{IEEEkeywords}
Image restoration, CNN-transformer cooperation, prompt
\end{IEEEkeywords}

\section{Introduction}
\label{introduction}
\IEEEPARstart{I}{mage} restoration aims to obtain a high-quality image from a given corrupted correspondence, $e.g.$, blurry, rainy, noisy or hazy image.  Over the years, image restoration has found extensive application across various real world scenario, including  autopilot, medical imaging, and surveillance. 

Although many methods have achieved excellent performance, such as denoising~\cite{noiseGuo2019Cbdnet,noiselee2022apbsn,noisezhang2023real},  debluring~\cite{blurkong2023efficient,blurpan2023cascaded,blurpan2023deep},  deraining~\cite{rain9798773,rainchen2023learning,rainyang2023alternating}, dehazing~\cite{hazeli2017aod,hazeqin2020ffa,hazesong2022rethinking,hazezheng2023curricular}, etc., these methods either focus solely on the task at hand or fine-tune the model individually for each specific task. When the degradation task and even the degree of degradation change, these methods yield  unsatisfactory results. This poses challenges to adopt them in the real world scenario as there must be multiple models, which makes the calculation complicated. In addition, real world environments are intricate; for instance, autonomous vehicles may encounter both rainy and hazy conditions simultaneously, which makes the system have to constantly decide and switch between a series of image restoration algorithms. Hence, it is crucial to develop an all-in-one algorithm capable of effectively restoring images where the 
degradation type even  corruption ratio is unknown\footnote{Noticed, in this paper, the “unknown” refers to the unavailability of information regarding the types of degradation and corruption ratio, and the “multiple degradations” indicates that each given image in the dataset contains only one degradation, while the dataset as a whole encompasses multiple types of degradations.}.

\begin{figure}[htb] 
	\centering
	\includegraphics[width=0.5\textwidth]{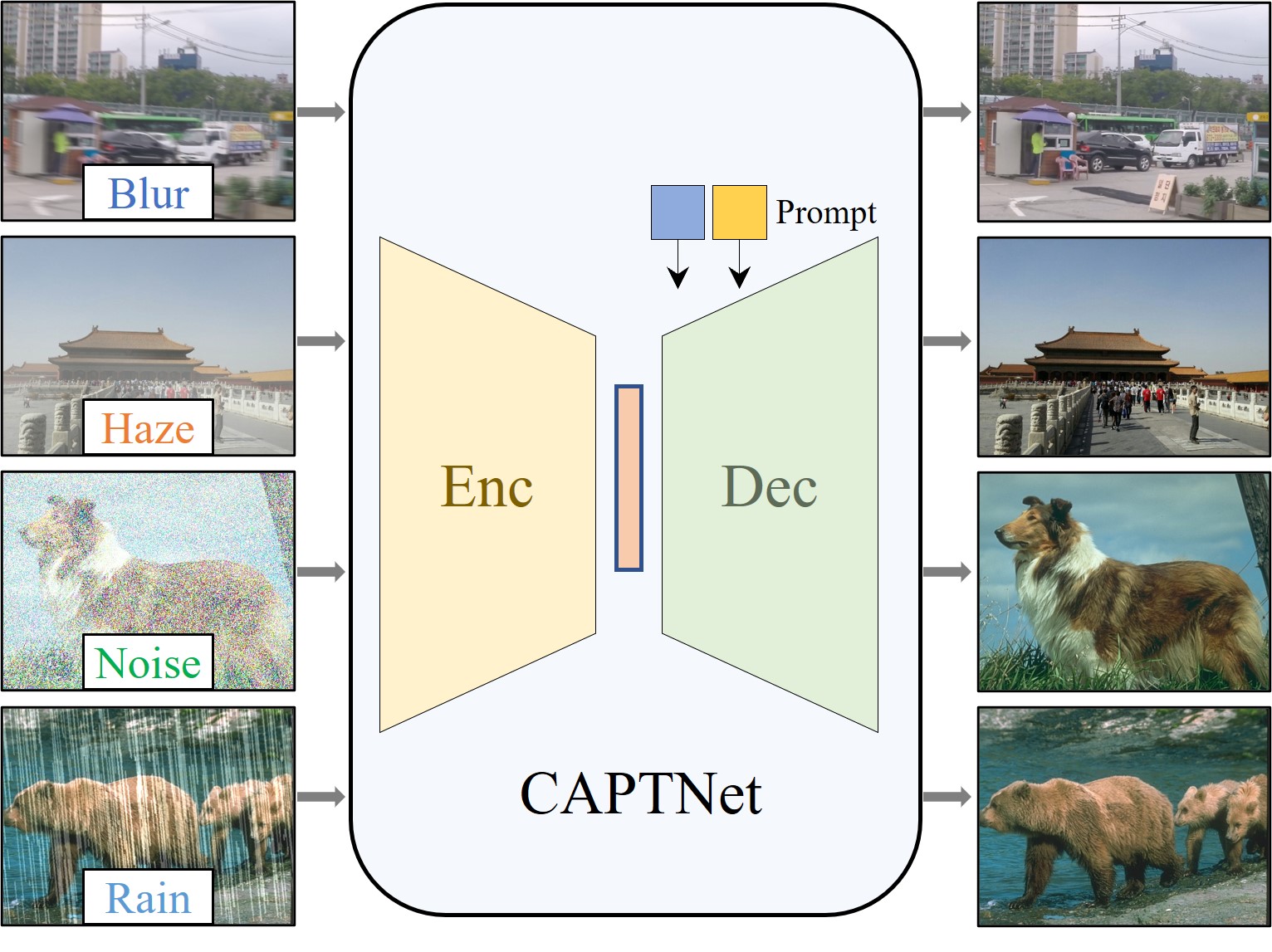}
	\caption{Illustrations of our basic idea. As shown, our proposed method, CAPTNet, has a single encoder and decoder, and inject a learnable prompt at multiple decoding stages to implicitly predict degradation conditions, which is used to guide the decoder to recover various degraded images adaptively.}
	\label{fig:01illustration}
\end{figure}

In recent times, there has been a notable emergence of all-in-one fashion methodologies, which effectively tackle multiple image degradations within a singular model. These methodologies can be broadly categorized into two distinct families, namely corruption-specific and corruption-agnostic. The former~\cite{IPT,li2020all} address various forms of degradation using individual subnetworks. However, this approach requires prior specification of the corruption types, which limits its broader applicability.  While the latter~\cite{valanarasu2022transweather,liu2022tape,airnetli2022all} overcome the constraints associated with the earlier corruption types,  enhancing the model's practical flexibility. However, the aforementioned models inadequately address the data-level connections associated with different degradation types, thereby failing to effectively handle the intricate interdependencies between various tasks. Moreover, as the number of tasks increases, their suffer from poor scalability. To solve the above problem,~\cite{IDRzhang2023ingredient} proposed a pioneering perspective to investigate degradation through the utilization of ingredient-oriented multi-degradation learning. Despite achieving remarkable results and possessing significant scalability, this innovative approach relies on tailored ad hoc operations for different degenerations based on the underlying physical principles. Additionally, it involves the pre-embedding of priors specific to distinct physics characteristics.

Based on the information presented, a natural question that comes to mind is whether it is feasible to design a end-to-end data ingredient-oriented approach to tackle multiple image degradation tasks within a single model? Thus, we propose a hypothesis that the target feature space of any image restoration task remains consistent (belonging to the domain of high-quality natural image features). Given specific prompts, the model can effectively handle various types of image degradation and adaptively map the feature space of the corrupted image to that of a high-quality image. To validate our hypothesis, we propose an integrated image restoration framework (as depicted  in Fig.~\ref{fig:01illustration}) that utilizes prompt learning in the encoder-decoder architecture, named as CAPTNet, which leverages prompt-based learning to guide adaptively recovering images affected by various degradations.
Specifically, the CAPTNet combines CNNs-based blocks and Transformer-based blocks to capture non-local information and local invariance. In the CNNs-based block, we did not make any innovation in the CNN-based block, but just used the nonlinear activation free block (NAFBlock)~\cite{chen2022simple}.  

While in the Transformer-based block, we design a simplified prompt-based transformer (SPT) architecture with  several key components. 1). We introduce a Multi-head Rearranged Attention with Prompts (MRAP). Prior to the attention calculation, the incorporation of local information is achieved through the implementation of a $1 \times 1$ convolution and a $3 \times 3$ depth-wise convolution. Subsequently, a rearrange operation is conducted to facilitate attention computation in the feature dimension rather than the spatial dimension. This adjustment ensures that the attention computation maintains a linear complexity based on the channel dimension.  Furthermore, we incorporate learnable prompt information into the $q, k,$  and $v$ matrices. This enables the model to effectively capture degenerate-type information and  guide the decoder in adaptively recovering images affected by various degradations.
2). We noted that the intensive computational pattern of self-attention makes feature interaction and aggregation processes vulnerable to implicit noise, leading to the consideration of uncorrelated representations in modeling global feature dependencies.  In order to produce enhanced features for the restoration of latent clear images, we have devised a  Simple Gate Feed-Forward Network (SGFN). This innovative approach incorporates a gating mechanism within the feed-forward network (FFN) to selectively determine the information that ought to be retained for the purpose of restoring latent clear images. Furthermore, we have eliminated the utilization of nonlinear activation functions, resulting in a reduction in computational expenses. Apart from the above architectural novelties, we incorporate a feature fusion module (FFM) further explores the multi-scale information  to improve the aggregated features.

The main contributions of this work are summarized below:
\begin{enumerate}
\item  We analyze  the data distribution characteristics of each image restoration task dataset, and propose a novel data ingredient-oriented approach that leverages prompt-based learning to enable a single model to efficiently tackle multiple image degradation tasks.
\item  We propose the CNNs and Prompt Transformer network (CAPTNet) combining CNNs operation and transformer to model the local invariant properties and non-local information for high-quality image restoration.
\item We  devise a Multi-Head Rearranged Attention with Prompts (MRAP) mechanism that efficiently combines local and non-local pixel interactions, guiding the model in adaptively restoring images impacted by diverse degradations through embedded learnable prompts.
\item We  design a effective Simple Gate Feed-Forward Network (SGFN) that selectively  determines which information should be preserved to facilitate the restoration of latent clear images.
\item We introduce a feature fusion module (FFM) to aggregate multi-scale features across encoder-decoder.
\item Extensive experiments demonstrate that our CAPTNet not only excels in adaptively addressing various degradation types, achieving state-of-the-art performance, but also competes  with task-specific algorithms.
\end {enumerate}
\begin{figure*}[htb] 
	\centering
	\includegraphics[width=1\textwidth]{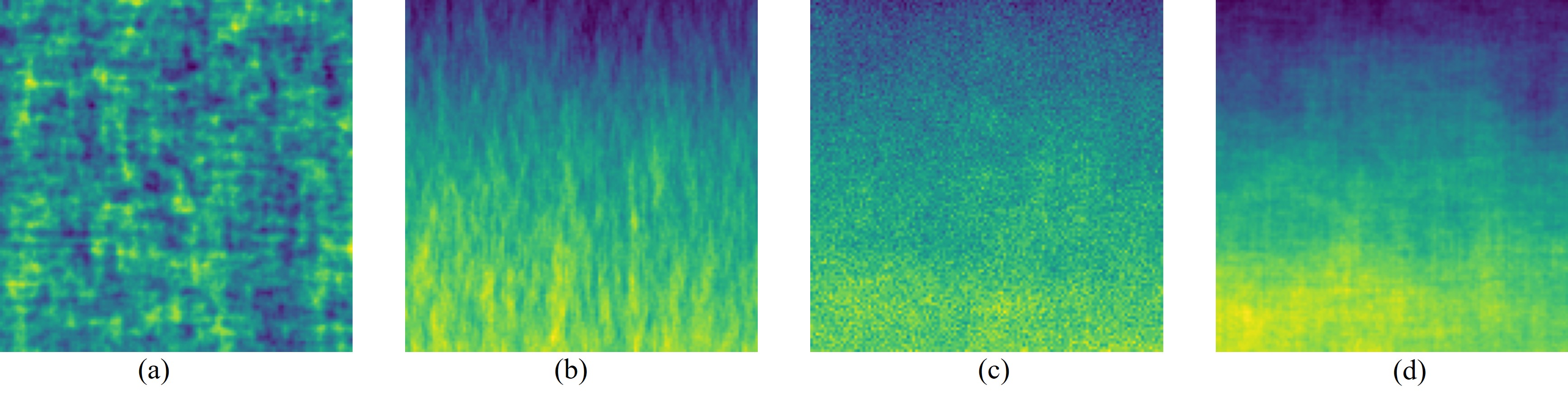}
	\caption{Visualized the differences between the clean image and its correspondence of individual image degradation datasets. (a) represents the GoPro~\cite{Gopro} dataset for image deblurring, (b) denotes the Rain100H~\cite{81Yang2016DeepJR} dataset for image deraining, (c) is the BSD300~\cite{BSDmartin2001database} for image denoising, and (d) shows the SOTS~\cite{RESIDEli2018benchmarking} dataset for image dehazing.}
	\label{fig:3diff_i_o}
\end{figure*}
\section{Related work}
\subsection{Image Restoration}
Image restoration tasks aim to restore a degraded image to a clean one, in that mitigating camera equipment or a variety of environmental factors during imaging. Early restoration approaches often relied on manually crafted priors to mitigate the ill-posed nature of the problems~\cite{1992Nonlinear, 1997Prior, 2005Fields, 2011Image, 2011Single}. In recent years, there has been a significant paradigm shift from traditional restoration methods to learning-based approaches, driven by their impressive performance in a wide range of image restoration tasks, such as denoising~\cite{noiseGuo2019Cbdnet,noiselee2022apbsn,noisezhang2023real}, dehazing~\cite{hazeqin2020ffa,hazeli2017aod,hazezheng2023curricular}, debluring~\cite{blurkong2023efficient,blurpan2023cascaded,blurpan2023deep}, deraining~\cite{rain9798773,rainchen2023learning,rainyang2023alternating}, etc. The learning-based approaches mentioned above can be classified into two main categories: CNN-based method and transformer-based method.
\subsubsection{CNNs Based Method}
Image restoration methods based on CNNs have made significant strides in recent years, with a plethora of emerging architectures~\cite{chen2022simple,cnnspairpurohit2021spatially,Zamir2022MIRNetv2,Zamir2021MPRNet,cnnmou2022deep,cnnpan2022dual}. \cite{Zamir2021MPRNet,cnnpan2022dual} pay attention to balance the competinng goals of
spatial details and high-level contextualized information. \cite{cnnspairpurohit2021spatially} leveraged the non-uniformity of severe degradations in the spatial domain, dynamically adapts computations to challenging regions within the image. \cite{Zamir2022MIRNetv2}  introduced a multi-scale architecture that effectively integrates contextual information while concurrently preserving  details. \cite{cnnmou2022deep} employed an expansion strategy to delve deeper into the principles of image restoration. \cite{chen2022simple} proposed a streamlined baseline network for image restoration, involving the removal or replacement of nonlinear activation functions. While these methods exhibit improved performance, they face challenges in capturing distant dependencies due to the constraints imposed by convolutional receptive fields. 

\subsubsection{Transformer Based Method}
With the flourish of vision transformers, their global modeling capability as well as the adaptability to input content have spawned a series of image restoration works. \cite{transchen2021preipt,transli2020learning} introduced  pre-training model based on transformers IPT and EDT respectively for image restoration tasks. \cite{transliang2021swinir} proposed SwinIR for image restoration based on Swin Transformer model. \cite{Zamir2021Restormer} devised a transposed attention mechanism and a feedforward neural network, culminating in the creation of an efficient transformer model. \cite{transWang_2022_CVPR_uform} proposed a U-Net based transformer architecture which uses local-enhanced window Transformer block.  Nevertheless, the transformer's capability to represent local invariant properties through self-attention is not as robust as that of CNNs, making it susceptible to confusion between background details and degraded information.

In this work, we combine CNNs and Transformers to capture long-dependent non-local information and model the local invariant properties.

\subsection{All-in-One Image Restoration}
All-in-one methods deal with multiple degradations within a single model. \cite{all_multie_li2020all} proposed a multi-encoder single-encoder to handle multiple bad weather degradations. \cite{transchen2021preipt} leveraged a multi-head, multi-tail architecture based on transformer to handle a variety of degradation. \cite{allvalanarasu2022transweather} employed weather-type queries to address various degradation issues through a single encoder-decoder transformer. \cite{all_conli2022all}  proposed a prior-free network with contrastive learning that does not differentiate different corruption types and ratios. \cite{all_liu2022tape} proposed to embed a task-agnostic prior into
a transformer. \cite{IDRzhang2023ingredient} introduced a novel perspective that delved into degradation through an ingredient-oriented approach, ultimately improving the model's scalability.

\subsection{Prompt learning}
Prompt learning methods performs a specific task or provides a response by processing and understanding natural language prompts or queries. Fueled by the birth of GPT-3~\cite{gptfloridi2020gpt}, it has achieved remarkable performance in a wide range of NLP tasks. In the field of visual domain, prompt learning is also employed to enhance the performance of specific tasks through fine-tuning. \cite{vptjia2022visual} introduced Visual Prompt Tuning (VPT) as an efficient and effective alternative to full fine-tuning for large-scale Transformer models.~\cite{pronie2022pro} proposed the method of parameter-efficient prompt tuning to enable the adaptation of frozen vision models to a wide range of downstream vision tasks. \cite{lionwang2023lion} proposed an implicit vision prompt tuning  model for various complex tasks with stable memory costs. All these approaches target high-level vision problems where image restoration tasks using prompt learning have not received attention. In this work, we propose an all-in-one model based on prompt learning, which guides the decoder to perform adaptive tackle multiple image degradation tasks according to the difference of data ingredient.

\section{Method}
\label{method}
\subsection{Problem Formulation and Motivation}
Image restoration aspires to restore a high-quality image from the low-quality one. the process of image degradation is physically formulated as
\begin{equation}
	\label{equ:01general}
	L = \mathbf{F}(\mathbf{H}) + N
\end{equation}
where $F(\cdot)$ indicates the degradation operator,  $N$ represents the additive noise, $L$ and $H$denotes an observed low-quality image and its corresponding high-quality image respectively. This formulation can signify different image restoration tasks when $F(\cdot)$ varies. If $F(\cdot)$ is the element-wise addition, it correspond to the denoising and deraining.  If $F(\cdot)$ is the element-wise addition, it correspond to the denoising and deraining.  If $F(\cdot)$ is the element-wise multiplication, it correspond to the dehazing. And if $F(\cdot)$ is a convolution operator, it correspond to the deblurring.
We visulaize the  differences between the clean image and its correspondence of individual image degradation datasets (See Fig.~\ref{fig:3diff_i_o}). It is evident that both in terms of data characteristics and problem definitions, these image restoration tasks do not share the same intrinsic characteristics, which explains why a task-special algorithm  does not work on the other tasks.

\begin{figure*}[htb] 
	\centering
	\includegraphics[width=1\textwidth]{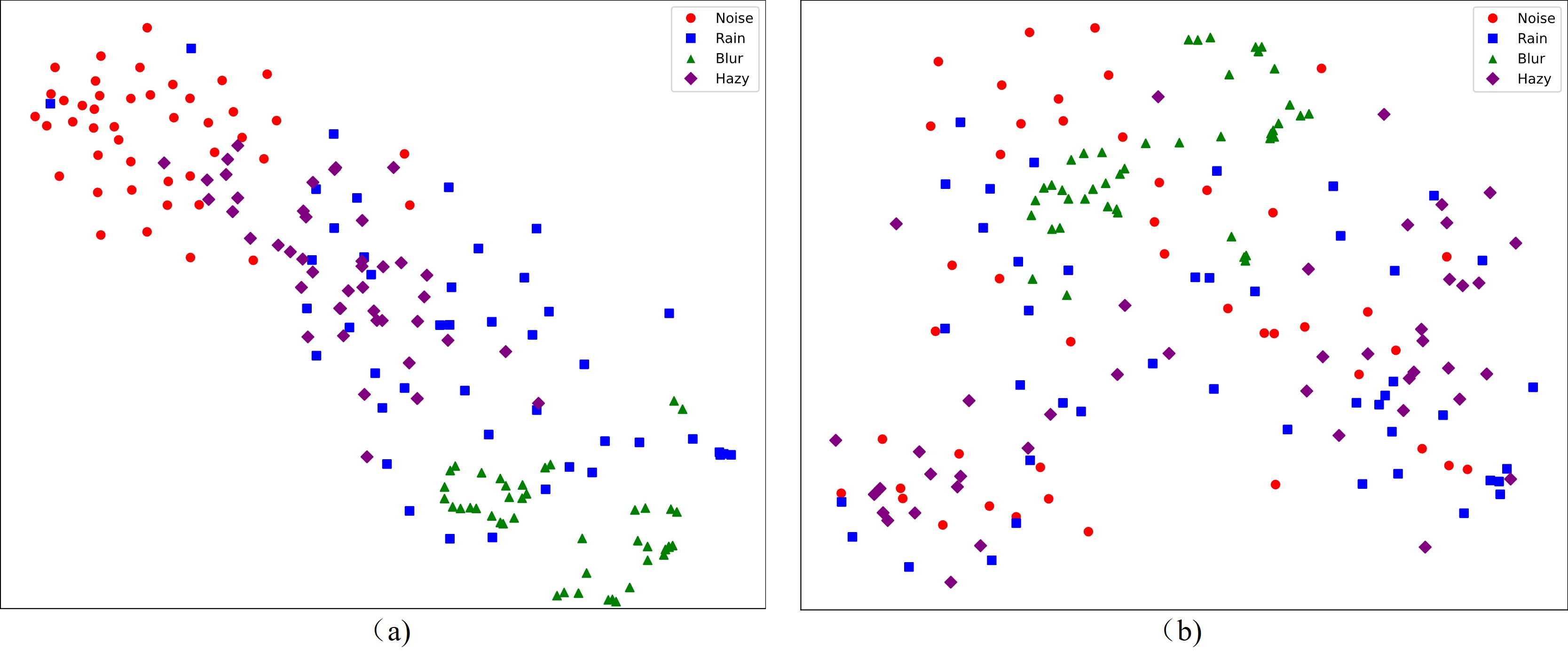}
	\caption{t-SNE visualization the distribution of data between datasets of individual image recovery tasks. Distinct colors denote different degradation types. (a) represents the corrupted image data (input data in the dataset), and (b) represents the corresponding high-quality image data (target data in the datasets). 
}
	\label{fig:02tsne_dataset}
\end{figure*}

  
However, what we have discovered is that despite these image restoration tasks originating from different corrupted image, they ultimately produce high-quality images. Based on this finding,  we analyzed the datasets of each image restoration task and visualized their data distribution (as shown in Fig.~\ref{fig:02tsne_dataset}). Fig.~\ref{fig:02tsne_dataset}(a) clearly illustrates that each image restoration task's dataset input possesses distinct characteristics. In particular, data with the same degradation type exhibit a notable proximity and clustering within the feature space.  Nevertheless, Fig.~\ref{fig:02tsne_dataset}(b) presents a scenario where the target image data from different degradation types are intricately intertwined within the feature space, rendering the detection of clear clustering phenomena a challenging task. We hypothesis that the target images within the dataset, encompassing various types of image degradation, are of high-quality, having been stripped of the degradation information specific to their respective types. Thus, they can be regarded as pristine natural images. When employing t-SNE to visualiz them, one is essentially engaging in a form of image classification for natural images.

\begin{figure}[htb] 
	\centering
	\includegraphics[width=0.5\textwidth]{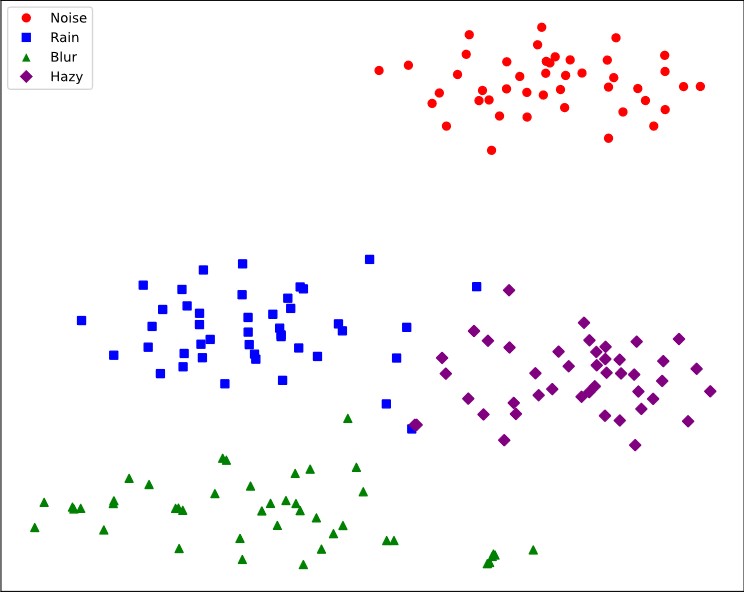}
	\caption{t-SNE visualization the output feature of encoder for each degradation type of data.}
	\label{fig:4capt_middle}
\end{figure}

This prompted us to consider whether we could "instruct" the network to identify and address specific types of image degradation, allowing it to adapt its restoration capabilities accordingly. With this objective in mind, we introduce CAPTNet, a comprehensive image restoration framework that incorporates prompt learning within the encoder-decoder architecture. This approach harnesses prompt-based learning to intelligently guide the adaptive recovery of images affected by a range of degradations. We formulat the image restoration tasks using the following universal function:
\begin{equation}
	\label{equ:01general_solution}
	 H = CAPTNet(L_t) = D_p \odot  E(L_t)
\end{equation}
where $L_t$  represent a degraded image with respect to a degradation type $t$, $D_p$ denotes  a decoder with a prompt that uses learnable prompt parameters $p$ to guide the model to adaptively recover images of various degraded types, and $E$ is the generic encoder that capture key features or information about the degraded image. It's important to highlight that the output feature map of encoder $E$ still retain information pertaining to their respective degradation types as shown in Fig.~\ref{fig:4capt_middle}. 

\begin{figure*}[htb] 
	\centering
	\includegraphics[width=1\textwidth]{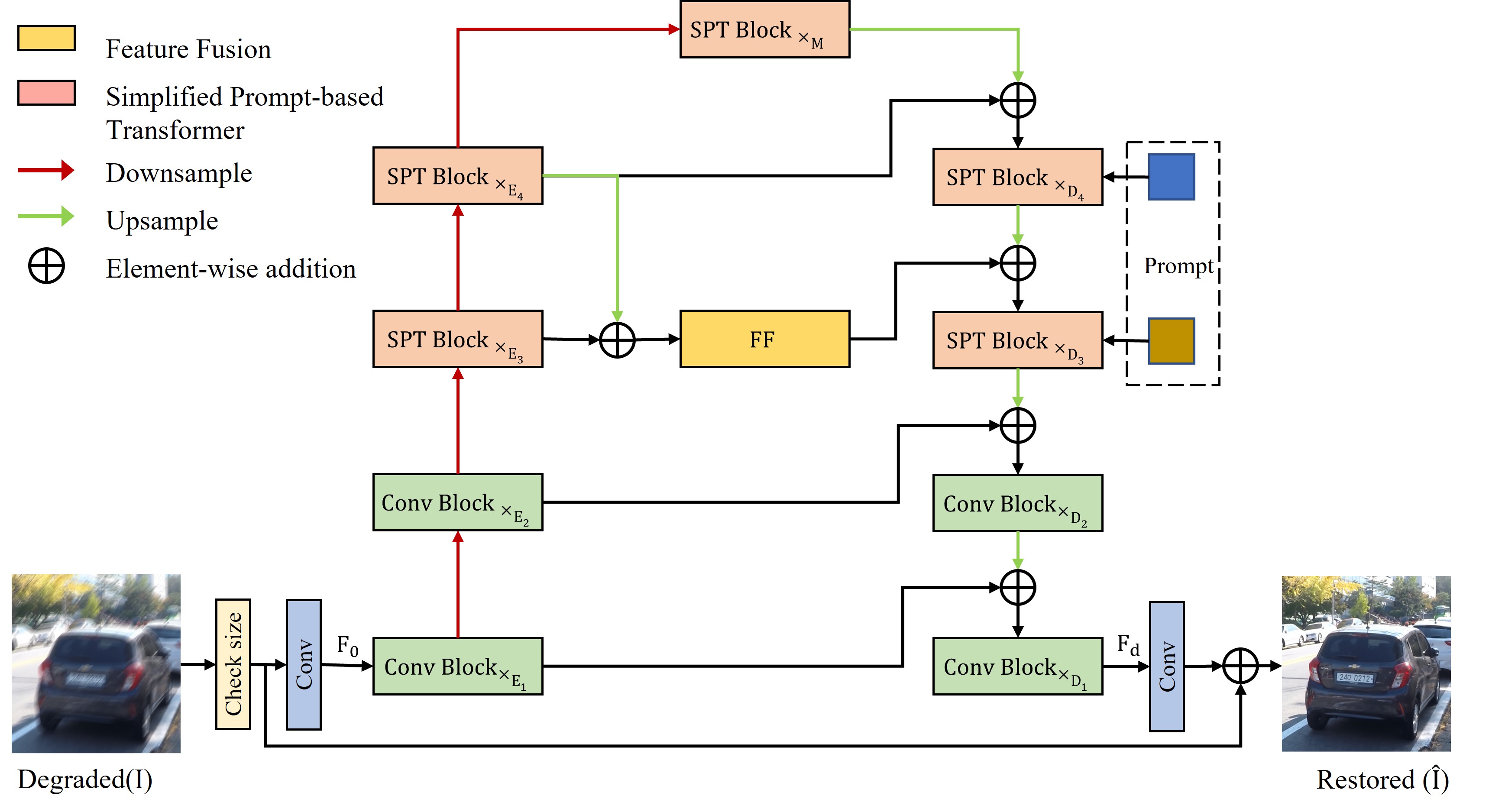}
	\caption{Architecture of CAPTNet for all-in-one image restoration. Our CAPTNet combines CNNs-based blocks (Conv Block) and Transformer-based blocks (SPT Block) to capture non-local information and local invariance. The details of Conv blocks and SPT blocks are shown in Fig.~\ref{fig:6component}. }
	\label{fig:5network}
\end{figure*}

\begin{figure*}[htb] 
	\centering
	\includegraphics[width=1\textwidth]{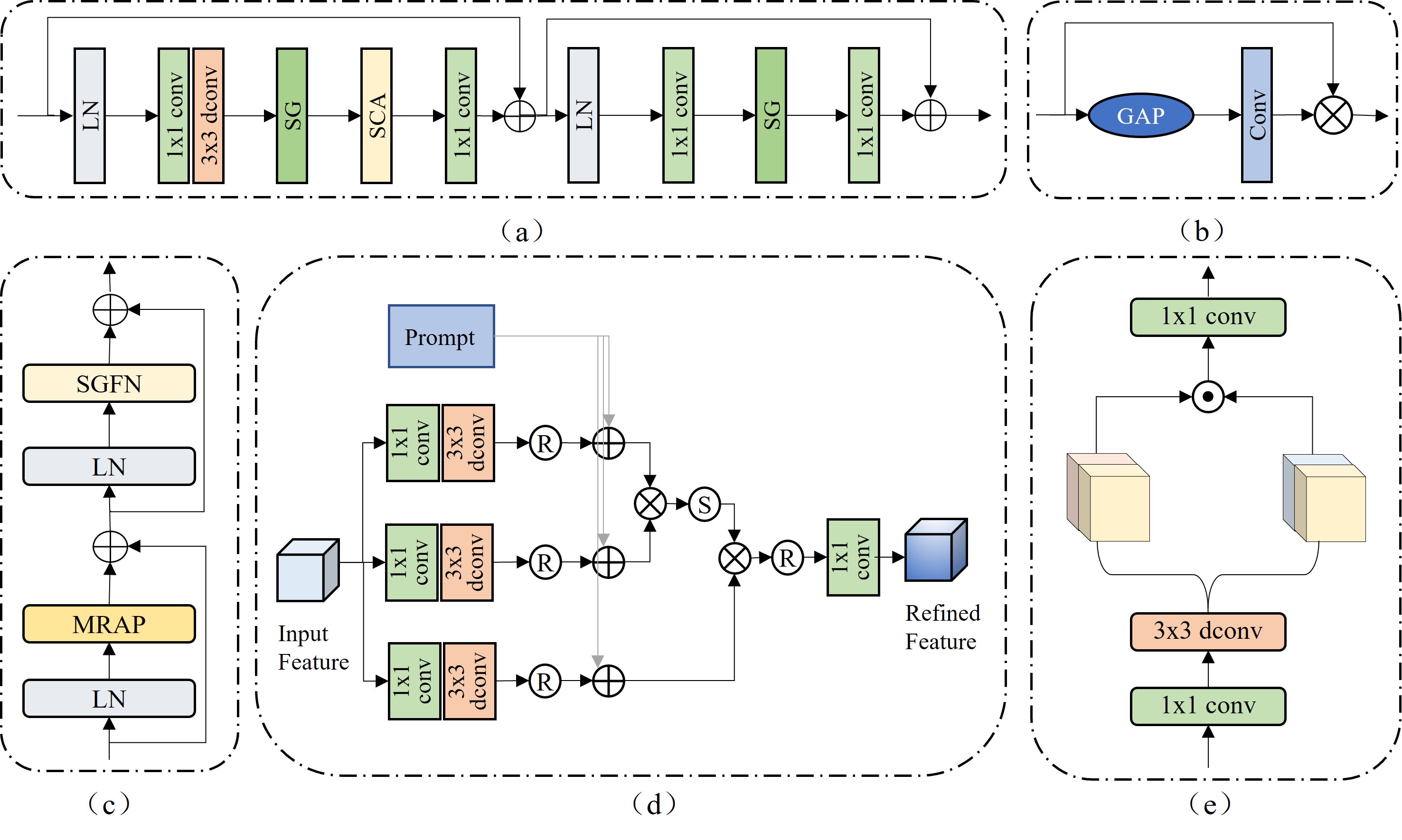}
	\caption{(a) The architecture of non-linear activation free block (NAFBlock)~\cite{chen2022simple}. (b) Simplified channel attention (SCA). (c) Simplified prompt-based transformer (SPT) block. (d) Multi-head rearranged attention with prompts (MRAP).  (e) Simple gate feed-forward network (SGFN)}
	\label{fig:6component}
\end{figure*}

\subsection{Network Architecture}
Our primary objective is to create a prompt-based learning network capable of performing all-in-one image restoration.  With this goal in mind, we present an integrated image restoration architecture CAPTNet (see Fig.~\ref{fig:5network}) that combines CNNs-based blocks (Conv Block) and simplified prompt-based transformer blocks (SPT Block). This integration effectively utilizes prompt-based learning to intelligently guide the adaptive recovery of images affected by a range of degradation types. We first present the overall pipeline of our CAPTNet architecture. Then we provide detailed descriptions of both the Conv block and the SPT block. The proposed SPT block comprises two fundamental components: (a) Multi-Head Rearranged Attention with Prompts (MRAP) and (b) Simple-Gate Feed-Forward Network (SGFN). Finally, we describe the  feature fusion module (FFM)  that aggregates multi-scale information.

\noindent\textbf{Overall Pipeline.} Given a degraded image $\mathbf{I} \in \mathbb R^{H \times W \times 3}$, CAPTNet first applies a $3 \times 3$ convolutional layer to extract shallow feature maps $\mathbf{F_{0}} \in \mathbb R^{H \times W \times C}$ ($H, W, C$ are the feature map height, width, and channel number, respectively). Next these shallow features $\mathbf{F_{0}}$ pass through $5-level$ encoder-decoder , yielding deep features $\mathbf{F_{D}} \in \mathbb R^{H \times W \times C}$. For feature down-sampling and up-sampling, we apply pixel-unshuffle and pixel-shuffle operations. To assist the revovery process, we add skip-connections to bridge across continuous intermediate features \cite{Zamir2021Restormer,Zamir2021MPRNet,transWang_2022_CVPR_uform}. While, we also add a FFM to integrate multi-scale information. To facilitate the decoding process, we introduce prompts into our CAPTNet framework. At every SPT in decoder level, the prompt block effectively enhances the input features by incorporating information regarding the degradation type, thereby guiding the recovery process. Finally, we apply convolution to deep  features $\mathbf{F_{D}}$ and generate a residual image $\mathbf{R}\in  R^{H \times W \times 3}$ to which degraded image is added to obtain the restored image: $\mathbf{\hat{I}} = \mathbf{R} +\mathbf{I}$. We optimize the proposed network using PSNR loss : 
\begin{equation}
	\label{equ:01}
	PSNR = 10 \cdot log_{10} \cdot \frac{(2^n-1)^2}{||\mathbf{\hat{I}}-\mathbf{\dot I}||^2}
\end{equation}

where $\mathbf{\dot I}$ denotes the ground-truth image. Next, we present the core of the CAPTNet.

\subsubsection{\textbf{Conv Block}}

As previously mentioned in our introduction, we have not introduced any innovations in the Conv block, but just used the nonlinear activation free block (NAFBlock)~\cite{chen2022simple}. Fig.~\ref{fig:6component}(a) illustrates the process of obtaining an output$Y$ from an input $X$ using Layer Normalization (LN), Convolution, Simple Gate (SG), and Simplified Channel Attention (SCA). Express as follows:
\begin{equation}
\begin{aligned}
	\label{equ:0xnaf}
	X_1 &= X + W_p^2(SCA(SG(W_d^1W_p^1(LN(X))))), \\
    Y &= X_1 + W_p^4(SG(W_p^3(LN(X_1)))), \\
    SG &= X_{f1} \cdot X_{f2}, \\
    SCA &= X_{f3} \cdot W_p^5 GAP(X_{f3})
\end{aligned}
\end{equation}
where $W_p^{(\cdot)}$ is the $1 \times 1$ convolution, $W_d^{(\cdot)}$ is the $3 \times 3$ depth-wise convolution, GAP is the global average pool,  and $X_{f1}, X_{f2} \in \  R^{H \times W \times \frac{C}{2}}$ are obtained by dividing $X_{f0}$ into channel dimensions. For a more intuitive presentation, we show $SCA(\cdot)$  in Fig.~\ref{fig:6component}(b). 

\subsubsection{\textbf{Simplified Prompt-based Transformer Block}}
There are two main challenges to apply standard Transformer~\cite{vaswani2017attention,vitdosovitskiy2020image} for image restoration. First, it compute self-attention globally across all tokens not only incurs additional computation cost relative to the number of tokens but also introduces noisy interactions between unrelated features, making it unfriendly for image restoration. Second, it  cannot adaptively recover images of different degradation types. To solve such limitations, we design a \textbf{S}implified \textbf{P}rompt-based \textbf{T}ransformer  (SPT)  as shown in Fig.~\ref{fig:6component}(c), which  selectively determines what information should be retained to facilitate efficient recovery of clear images, but also guides models to adaptively recover images affected by various degradations through embedded learnable prompts.
Formally, given the input features at the $(l-1)_{th}$ block $X_{l-1}$, the computation of a SPT block is represented as:
\begin{equation}
\begin{aligned}
	\label{equ:08spt}
   X_{l}^1 &= X_{l-1} + MRAP(LN(X_{l-1})),
\\
   X_{l} &= X_{l}^1 + SGFN(LN(X_{l}^1)) 
\end{aligned}
\end{equation}
where LN denotes the layer normalization; $X_{l}^1$ and $X_{l}$ denote the outputs from the  multi-head rearranged attention with prompts (MRAP) and simple gate feed-forward network (SGFN). In the following, we  describe MRAP and SGFN in details.

\noindent\textbf{Multi-head Rearranged Attention with Prompts.}
The vanilla self-attention (SA) paradigm requires computing the attention map for all tokens, the computation on a global scale results in a quadratic complexity:
\begin{equation}
	\label{equ:10complex}
    \mathcal{O}_{SA} = 4HWC^2 + 2(HW)^2C  
\end{equation}
Hence, applying SA to most image restoration tasks, which typically involve high-resolution images, is impractical and not feasible. In our work, we develop MRAP, shown in Fig.~\ref{fig:6component}(d), that introduces prompt information and has linear complexity. Specifically, given the input feature $\textbf{F}$ with a spatial resolution of $H \times W$ and $ C$ channels, we first apply $1 \times 1$ convolutions and $3 \times 3$  depth-wise convolutions to aggregate channel-wise context, yielding $ \textbf{Q} = W_d^QW_p^Q\textbf{F}$, $\textbf{K} = W_d^KW_p^K\textbf{F}$, $\textbf{V} = W_d^VW_p^V\textbf{F}$. Next, we rearrange $query, key$ and $value$ matrices $\mathbf{Q} \in \mathbb R^{H \times W \times C}$, $\mathbf{K} \in \mathbb R^{H \times W \times C}$ and $\mathbf{V} \in \mathbb R^{H \times W \times C}$  to $\mathbf{\hat{Q}} \in \mathbb R^{(H W) \times \frac{C}{h} \times h}$, $\mathbf{\hat{K}} \in \mathbb R^{(H W) \times \frac{C}{h} \times h}$, $\mathbf{\hat{V}} \in \mathbb R^{(H W) \times \frac{C}{h} \times h}$, where $h$ is the number of head. Then we inject learnable promopts to make the model better understand the degradation types as follows:
\begin{equation}
\begin{aligned}
	\label{equ:10injectp}
     \Tilde{Q} &= \hat{Q} + P_q
     \\
     \Tilde{K} &= \hat{K} + P_k
     \\
     \Tilde{V} &= \hat{V} + P_v
\end{aligned}
\end{equation}
Inspired by~\cite{Zamir2021Restormer}, we apply SA across channels rather than spatial dimensions to reduce the computation complexity. The attention matrix is thus computed by the self-attention mechanism  as :
\begin{equation}
	\label{equ:10mrap}
    Attention(\Tilde{Q}, \Tilde{K}, \Tilde{V}) =  Softmax(\frac{\Tilde{Q}\Tilde{K}}{\beta}) \Tilde{V}        
\end{equation}
where $\beta$ is a learning scaling parameter used to adjust the magnitude of the dot product of $\Tilde{Q}$ and $\Tilde{K}$ prior to the application of the softmax function. Finally, we rearrange the attention matrix back to its original dimensions of $\mathbb R^{H \times W \times C}$,  and apply a $1\times 1$ convolutions to generate the output feature \textbf{$\hat{F}$}. After the above process, the computation change from  quadratic complexity $\mathcal{O}_{SA}$ to  linear complexity $\mathcal{O}_{MRAP}$.
\begin{equation}
	\label{equ:101mrapcomplex}
    \mathcal{O}_{MRAP} = 5hwC^2 + hwC
\end{equation}

\noindent\textbf{Simple Gate Feed-Forward Network.}
Previous studies~\cite{Zamir2021Restormer,transWang_2022_CVPR_uform} usually add depth-wise convolutions to improve the capability to leverage local context. However, those exploitations all ignore the implicit noise that contains in the output from intensive computational of self-attention. In this work,  we propose a \textbf{S}imple \textbf{G}ate \textbf{F}eed-Forward \textbf{N}etwork (SGFN) that directly divide the feature map into two parts in the channel dimension and selectively  determine which frequency information should be preserved for restoring latent clear images. Given an input tensor $\textbf{T} \in \mathbb R^{H \times W \times C}$ The proposed SGFN can be formulated by:
\begin{equation}
\begin{aligned}
	\label{equ:0032sgfn}
     \mathbf{\hat{T}} &= W_p^0 SGating(\textbf{T})
     \\
     SGating((\textbf{T})) &= SG(W_d^1 W_p^1 \textbf{T})
\end{aligned}
\end{equation}
where $SG(\cdot)$ represents the simple gate that has been formulated in Eq.~\ref{equ:0xnaf}. Noted that we remove the nonlinear activation function which reduces the computational complexity.  The detailed network architecture of the proposed SGFN is shown in Fig.~\ref{fig:6component}(e).

\begin{figure}[htb] 
	\centering
	\includegraphics[width=0.4\textwidth]{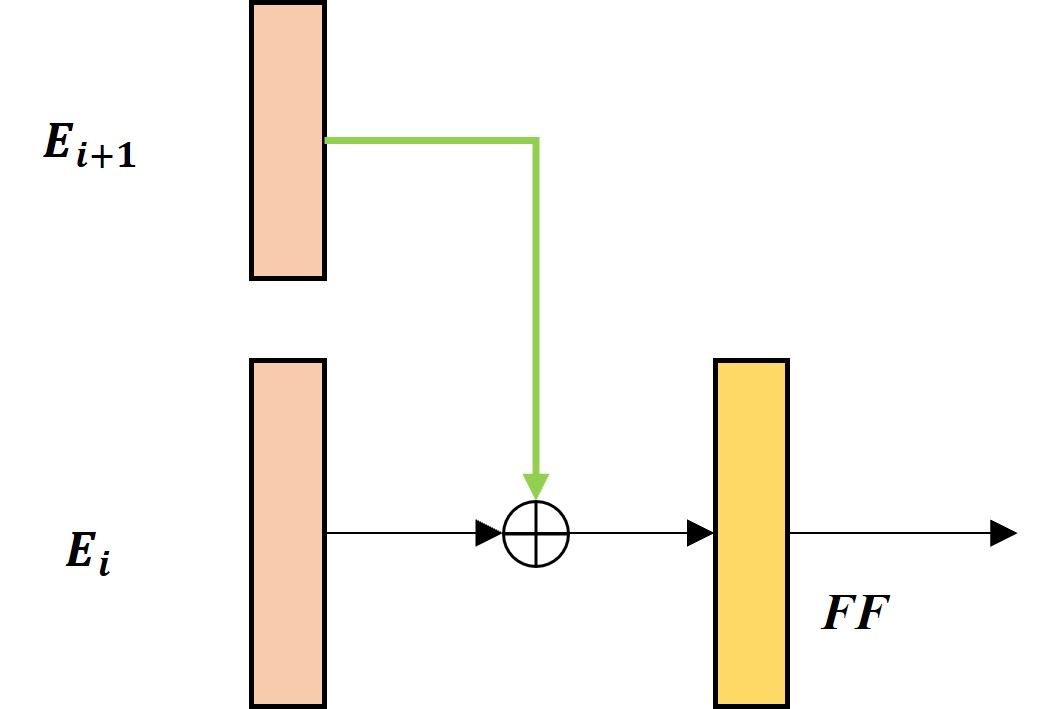}
	\caption{The proposed feature fusion module (FFM) that enables the exchange of information
across multiple scales. }
	\label{fig:0011ffm}
\end{figure}
\subsubsection{\textbf{Feature Fusion Module}}

In our framework, we introduce the \textbf{F}eature \textbf{F}usion \textbf{M}odule (FFM) to fuse the information of levle-3 and level-4 SPT encoders (see Fig.~\ref{fig:0011ffm}). After upsampling the level-4 SPT encoder output feature map \textbf{$E_4$}, we seamlessly integrate it with the level-3 SPT encoder output feature map \textbf{$ E_3$} through element-wise addition. The resulting combined feature map is then fed into FF to generate the final fusion feature \textbf{$F_o$}.  The entire feature fusion procedure of the developed FFM is formulated as:
\begin{equation}
	\label{equ:012ffm}
	\mathbf{F_o} = FF(Up(\mathbf{E_4}) \oplus \mathbf{E_3})
\end{equation}
where $\oplus$ denote the element-wise addition, $UP(\cdot)$ represents the up-sampling operation and $FF(\cdot)$ is the feature fusion operate which we use a layer of NAFBlock (see Fig.~\ref{fig:6component}(a))~\cite{chen2022simple}. This design offers two benefits. Firstly, it integrates multi-scale information, allowing the network model to capture abundant context information. Second, it enriches the features and is more conducive to decoder adaptive recognition of image degradation types.

\section{Experiments}
\label{exper}
In this section, we evaluate the proposed method under two different experimental settings: (a) All-in-One, and (b) Task-Special. In the following, we will first introduce the experimental setting and  then present the qualitative and quantitative comparison results on benchmarks. Finally, we will conduct some ablation studies to verify the effectiveness of our method. 

\subsection{Experimental Settings}
In this section, we introduce the details of the used datasets, metrics, and training details.

\begin{table*}
\caption{ Dataset description for various image restoration tasks.}
\label{tb:datadec}
    \centering
    \begin{tabular}{c|c|c|c|c|c}
         \hline
         Tasks & Datasets & Train Samples & Total Training Sample &Test Samples &Testset Rename
          \\
        \hline
          \multirow{7}{*}{Deraining} & Rain14000~\cite{238099669} &11200 &\multirow{7}{*}{13712} &0 & -
         \\
         &Rain1800~\cite{81Yang2016DeepJR} &1800 & &0 & -
            \\
         &Rain12~\cite{487780668} &12 & &0 &- 
        \\
         &Rain800~\cite{90Zhang2017ImageDU} &700 & & 98 &Test100
        \\
         &Rain1200~\cite{DIDMDN} &0 & &1200 &Test1200
        \\
         &Rain100H~\cite{81Yang2016DeepJR} &0 & &100 & Rain100H
        \\
         &Rain100L~\cite{81Yang2016DeepJR} &0 & &100 &Rain100L
         \\
         \hline
         \multirow{2}{*}{Deblurring} & GoPro~\cite{Gopro} &2130 & \multirow{2}{*} {2130}&1111 & GoPro
         \\
         &HIDE~\cite{HIDE} &0 & &2025 &HIDE
         \\
         \hline
         \multirow{3}{*}{Denoising} & WED~\cite{wedma2016waterloo} & 4744 &\multirow{3}{*} {4944} & 0 & -
         \\
          & BSD300~\cite{BSDmartin2001database} & 200 && 0 & -
         \\
         & BSD68~\cite{BSDmartin2001database} & 0 && 68 & BSD68
         \\
         \hline
         \multirow{2}{*}{Dehazing} & RESIDE~\cite{RESIDEli2018benchmarking} & 13990 &\multirow{2}{*} {13990}& 0 & -
         \\
         & SOTS-Outdoor~\cite{RESIDEli2018benchmarking} & 0 & & 492 & SOTS
         \\
         \hline
    \end{tabular}
\end{table*}
\subsubsection{\textbf{Datasets}}

The datasets used for training and testing are summarized in Table.~\ref{tb:datadec}. Next, we describe the datasets used for (1) Task-Special and (2) All-in-One.

\noindent\textbf{Task-Special.} Since our approach is primarily concerned with the ability to achieve all-in-one image restoration, it has only been demonstrated in image deraining and image deblurring that our approach can achieve impressive results as a task-special model.
For image deraining, we train our model on 13,712 clean-rain image pairs gathered from multiple datasets~\cite{81Yang2016DeepJR,487780668,90Zhang2017ImageDU,238099669}, and perform evaluation on Rain100L~\cite{81Yang2016DeepJR}, Rain100H~\cite{81Yang2016DeepJR}, Test100~\cite{90Zhang2017ImageDU} and Test1200~\cite{DIDMDN}. For image deblurring, we utilize the GoPro~\cite{Gopro} dataset, which consists of 2,103 image pairs for training and 1,111 pairs for evaluation. Additionally, we assess the generalizability of our approach by applying the GoPro-trained model directly to the test images of the HIDE~\cite{HIDE} dataset which comprises 2,025 images.

\begin{figure}[htb] 
	\centering
	\includegraphics[width=0.5\textwidth]{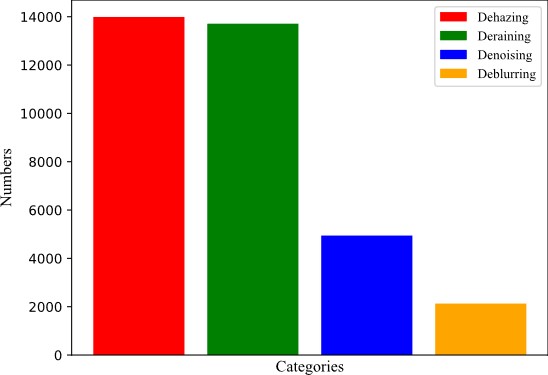}
	\caption{The number of datasets for each type of image restoration task.}
	\label{fig:data_number}
\end{figure}
\noindent\textbf{All-in-One.} We train our method on a combination of multiple image degradation datasets. As shown in Table.~\ref{tb:datadec}, including Rain14000~\cite{238099669}, Rain1800~\cite{81Yang2016DeepJR}, Rain12~\cite{487780668} and Rain800~\cite{90Zhang2017ImageDU} for deraining, RESIDE~\cite{RESIDEli2018benchmarking} for dehazing, BSD300~\cite{BSDmartin2001database} and WED~\cite{wedma2016waterloo} for denoising and GoPro~\cite{Gopro} for deblurring. For evaluation, GoPro~\cite{Gopro}, Rain100L~\cite{81Yang2016DeepJR}, BSD68~\cite{BSDmartin2001database} and SOTS~\cite{RESIDEli2018benchmarking} are utilized as the test sets. It is worth noting that, as shown in Figure 1, there is a big difference in the number of datasets for each degradation task, such as 13990 for dehazing and 2130 for deblurring, which will make our model focus more on the tasks with large amounts of data during the training process. To this end, we perform data equalization operations so that the data sets for each degradation task tend to be similar.

\subsubsection{\textbf{ Metrics}}
We employ two popular metrics for quantitative comparisons: Peak Signal-to-Noise Ratio (PSNR) and Structural Similarity (SSIM). Higher values of these metrics indicate superior performance of the methods.
\subsubsection{\textbf{Training details}}
We first trained the deraining and deblurring tasks separately, and then used the pre-trained deblurring model to train the all-in-one task using the combination dataset. We utilize the following block configurations in our network for each level: $[1,1,1,28]$ blocks for the encoder, $[1,1,1,1]$ blocks for the decoder. For task-special,  we train models with Adam~\cite{2014Adam} optimizer($\beta_1=0.9, \beta_2=0.999$) and PSNR loss for $5 \times 10^5$ iterations with the initial learning rate $5 \times 10^{-4}$ gradually reduced to $1 \times 10^{-7}$  with the cosine annealing~\cite{2016SGDR}. We extract patches of size $256 \times 256$ from training images, and the batch size is set to $32$. For All-in-One, we set the initial learning rate $5 \times 10^{-5}$, all else unchanged.
We adopt TLC~\cite{Chu2021ImprovingIR} to solve the issue of performance degradation caused by training on patched images and testing on the full image. For data augmentation, we perform horizontal and vertical flips. 

\begin{table*}
  \caption{ Quantitative results under All-in-One restoration setting with state-of-the-arts general image restoration and all-in-one methods.  The best and the second best results are marked in \textbf{bold} and \underline{underlined}, respectively. When averaged across different tasks, our CAPTNet provides a significant gain of 0.32 dB over the previous all-in-one method IDR~\cite{IDRzhang2023ingredient}}
    \label{tab:experi_all_in_one}
    \centering
    \begin{tabular}{ccccccccc|cc}
    \\
    \hline
    \multicolumn{1}{c}{} & \multicolumn{2}{c}{GoPro~\cite{Gopro}}  & \multicolumn{2}{c}{Rain100L~\cite{81Yang2016DeepJR}} & \multicolumn{2}{c}{BSD68~\cite{BSDmartin2001database}} & \multicolumn{2}{c|}{SOTS~\cite{RESIDEli2018benchmarking}} & \multicolumn{2}{c}{Average} 
    \\
   Methods &PSNR $\uparrow$ &  SSIM $\uparrow$  & PSNR $\uparrow$ & SSIM $\uparrow$ &PSNR $\uparrow$ &SSIM $\uparrow$ & PSNR $\uparrow$&SSIM $\uparrow$ & PSNR $\uparrow$ & SSIM $\uparrow$  
    \\
    \hline
    HINet~\cite{Chen_2021_CVPRhinet} & 30.22 & 0.824 & 35.43 & 0.969 & 30.50 & 0.881 & 26.77 & 0.935 & 30.73 & 0.902
    \\
    MPRNet~\cite{Zamir2021MPRNet} & 31.97 & 0.934 & \underline{37.25} & \textbf{0.980} & \underline{31.03} & 0.889 & 26.24 & 0.936 & 30.97 & 0.935 
    \\
    NAFNet~\cite{chen2022simple} & 32.03 & 0.941 & 31.45 & 0.966 & 30.86 & 0.881 & 27.09 & 0.925 & 30.86 & 0.928
    \\
    MIRNetV2~\cite{Zamir2022MIRNetv2} & 31.55 & 0.903  &31.82 & 0.953 & 30.77 & 0.876 & 26.01 & 0.925 & 30.64 & 0.914
    \\
    SwinIR~\cite{transliang2021swinir} & 29.88 & 0.873 & 31.77 & 0.921 & 30.12 & 0.861 & 23.37 & 0.889 & 28.64 & 0.886
    \\
    Restormer~\cite{Zamir2021Restormer} & 32.32 & 0.948 & 36.66& 0.961 &31.01 & 0.874 & 27.83& 0.920 & 31.96 & 0.926
    \\
    \hline
    DL~\cite{fan2019general} & 23.67& 0.756 & 21.88 & 0.761 & 23.02 & 0.744 & 20.52 & 0.825 & 22.29 & 0.772
    \\
    Transweather~\cite{allvalanarasu2022transweather} & 30.82 & 0.867 & 34.43 & 0.905 & 29.80 & 0.841 & 25.32 & 0.885 & 29.02 & 0.875
    \\
    TAPE~\cite{all_liu2022tape} & 31.74 & 0.903 & 34.66 & 0.903 & 30.08 & 0.851 & 26.01 & 0.858 & 30.47 & 0.879
    \\
    AirNet~\cite{airnetli2022all} & 32.35 & 0.951 & 36.82 & 0.967 & 30.74 & 0.880 & 29.01 & 0.967 & 32.23 & 0.941
    \\
    IDR~\cite{IDRzhang2023ingredient} & \underline{32.37} & \underline{0.952} & 36.52 & 0.973 & \textbf{31.15} & \underline{0.886} & \textbf{29.33} & \textbf{0.972} & \underline{32.33} & \underline{0.946}
    \\
    \textbf{CAPTNet (Ours)} & \textbf{32.71} & \textbf{0.960} & \textbf{37.86} & \underline{0.977} & 30.75 & \textbf{0.896} & \underline{29.28} & \underline{0.968} & \textbf{32.65} & \textbf{0.950}
    \\
    \hline
    \end{tabular}
  
\end{table*}
\begin{figure*}[htb] 
	\centering
	\includegraphics[width=1\textwidth]{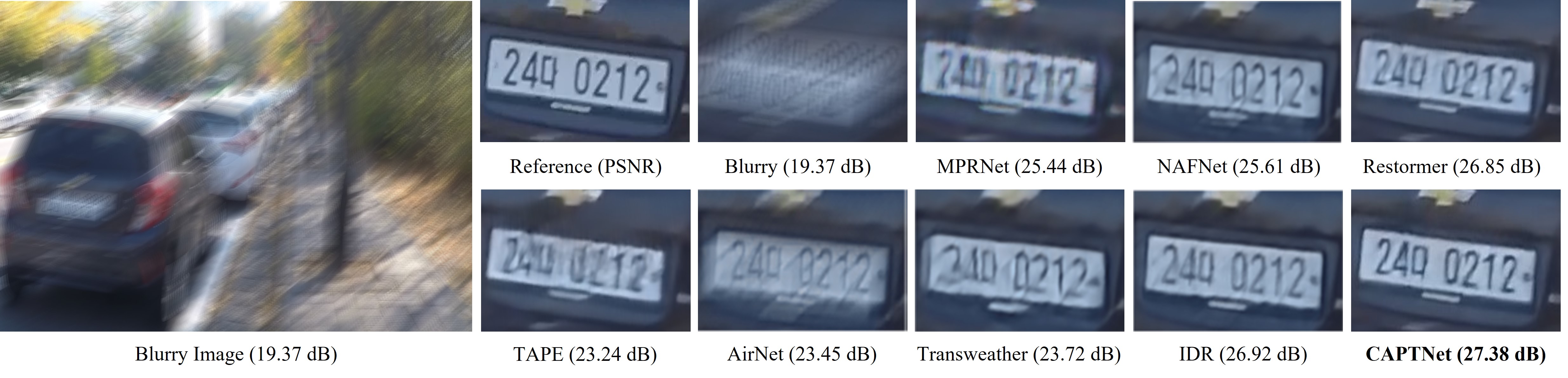}
	\caption{Visual under All-in-One restoration setting compare with state-of-the-art methods on GoPro dataset.}
	\label{fig:0blur_a}
\end{figure*}

\begin{figure*}[htb] 
	\centering
	\includegraphics[width=1\textwidth]{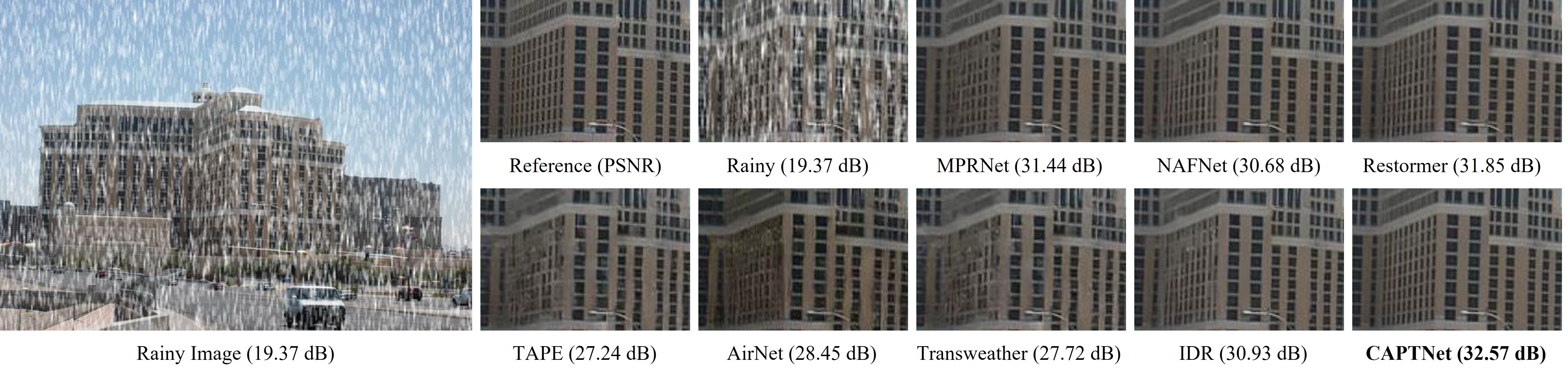}
	\caption{Visual under All-in-One restoration setting compare with state-of-the-art methods.}
	\label{fig:0rain_a}
\end{figure*}

\subsection{Comparison with state-of-the-art methods}
We compare our model with the advanced model in three experimental types: (1) Multiple Degradations All-in-One, (2) Single Degradation Task-Special and (3) Unkonwn tasks. The experimental results demonstrate that the proposed method serves as both an all-in-one model for adaptively restoring images with various types of degradation and a task-specific model. The results are described in detail below.

\subsubsection{\textbf{Multiple Degradations All-in-One Results}}
\begin{figure*}[htb] 
	\centering
	\includegraphics[width=1\textwidth]{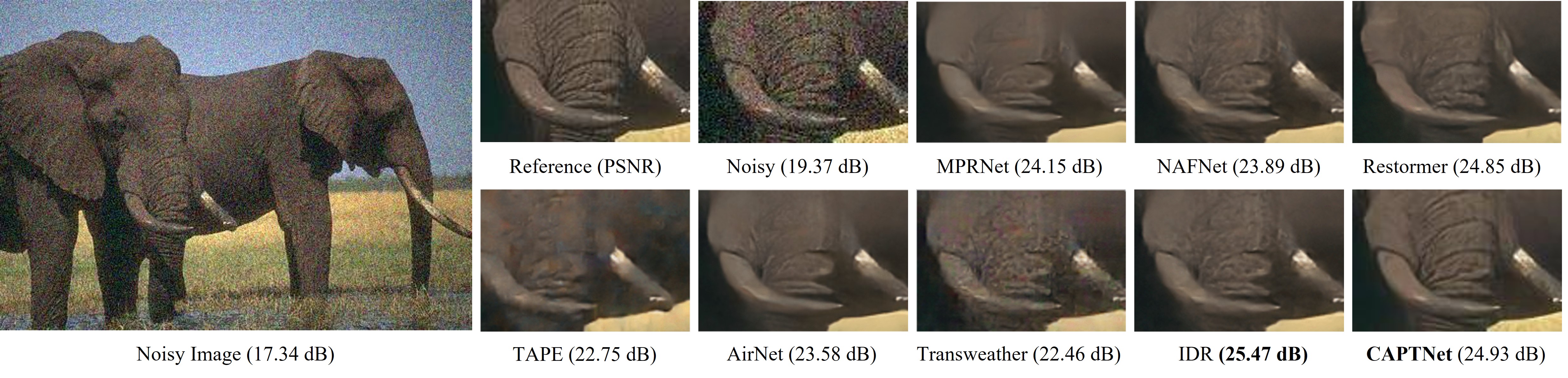}
	\caption{Visual under All-in-One restoration setting compare with state-of-the-art methods on BSD68 dataset. }
	\label{fig:0noise_a}
\end{figure*}

\begin{figure*}[htb] 
	\centering
	\includegraphics[width=1\textwidth]{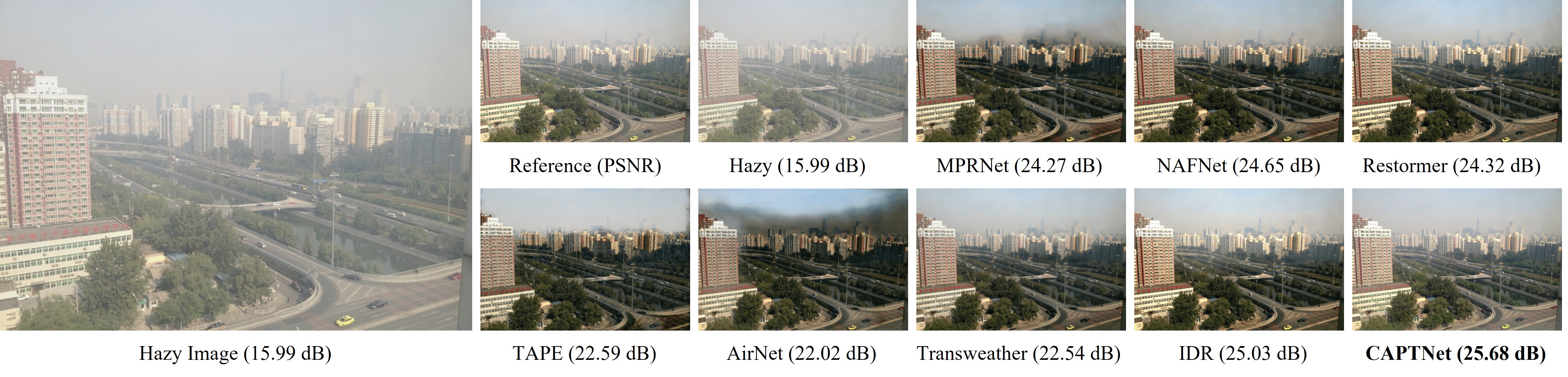}
	\caption{Visual under All-in-One restoration setting compare with state-of-the-art methods on SOTS dataset.}
	\label{fig:0haza_a}
\end{figure*}

We compare our CAPTNet with six general image restoration methods~\cite{Chen_2021_CVPRhinet,Zamir2021MPRNet,chen2022simple,Zamir2022MIRNetv2,transliang2021swinir,Zamir2021Restormer} and five all-in-one fashion methods~\cite{fan2019general,allvalanarasu2022transweather,all_liu2022tape,airnetli2022all,IDRzhang2023ingredient} on four challenging image restoration tasks including deblurring, deraining, denoising and dehazing. 
Table.~\ref{tab:experi_all_in_one} reports the quantitative comparison results.
When averaged across different restoration tasks, our algorithm yields 0.32 dB performance gain over the previous best method IDR~\cite{IDRzhang2023ingredient}.
Specifically, the proposed CAPTNet significantly advances state-of-the-art by providing 0.34 dB PSNR improvement on the image deblurring task. The visual examples provided in Fig.~\ref{fig:0blur_a} shows that the images restored by our model are sharper and closer to the ground-truth than those by others. Similarly on the image deraining task, the proposed CAPTNet achieves a substantial gain of 0.61 dB compared to MPRNet~\cite{Zamir2021MPRNet} and 1.04 dB over AirNet~\cite{airnetli2022all}. Visual comparisons in Fig.~\ref{fig:0rain_a} shows that CAPTNet is effective in removing rain streaks of different orientations and magnitudes, and generates images that are visually pleasant and faithful to the ground truth. On the denoising task, although our PSNR index is quite low, our SSIM score is the highest, surpassing the second highest by 0.01. This suggests that our method effectively preserves the structural information in the images.
Fig.~\ref{fig:0noise_a} illustrates visual results. Our method is able to remove real noise, while preserving the structural information.  Finally, on the dehazing task, although we are not the best, as shown in Fig.~\ref{fig:0haza_a} the image recovered by our model is much closer to ground truth.


\subsubsection{\textbf{ Single Degradation Task-Special Results}}
\begin{figure*}[htb] 
	\centering
	\includegraphics[width=1\textwidth]{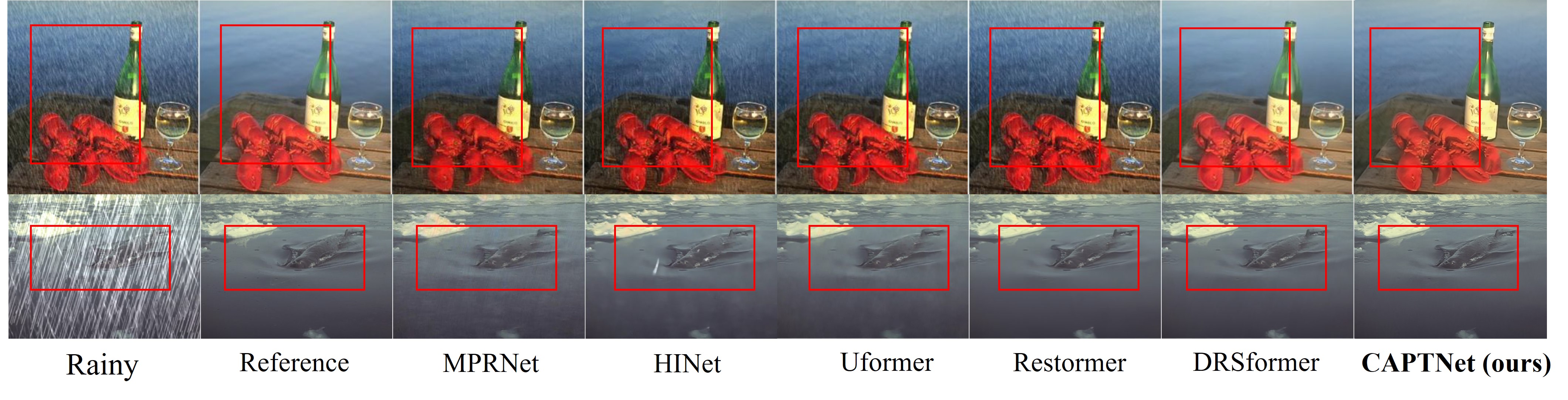}
	\caption{Visual under Task-special restoration setting compare with state-of-the-art methods}
	\label{fig:0rain_s}
\end{figure*}

\begin{figure*}[htb] 
	\centering
	\includegraphics[width=1\textwidth]{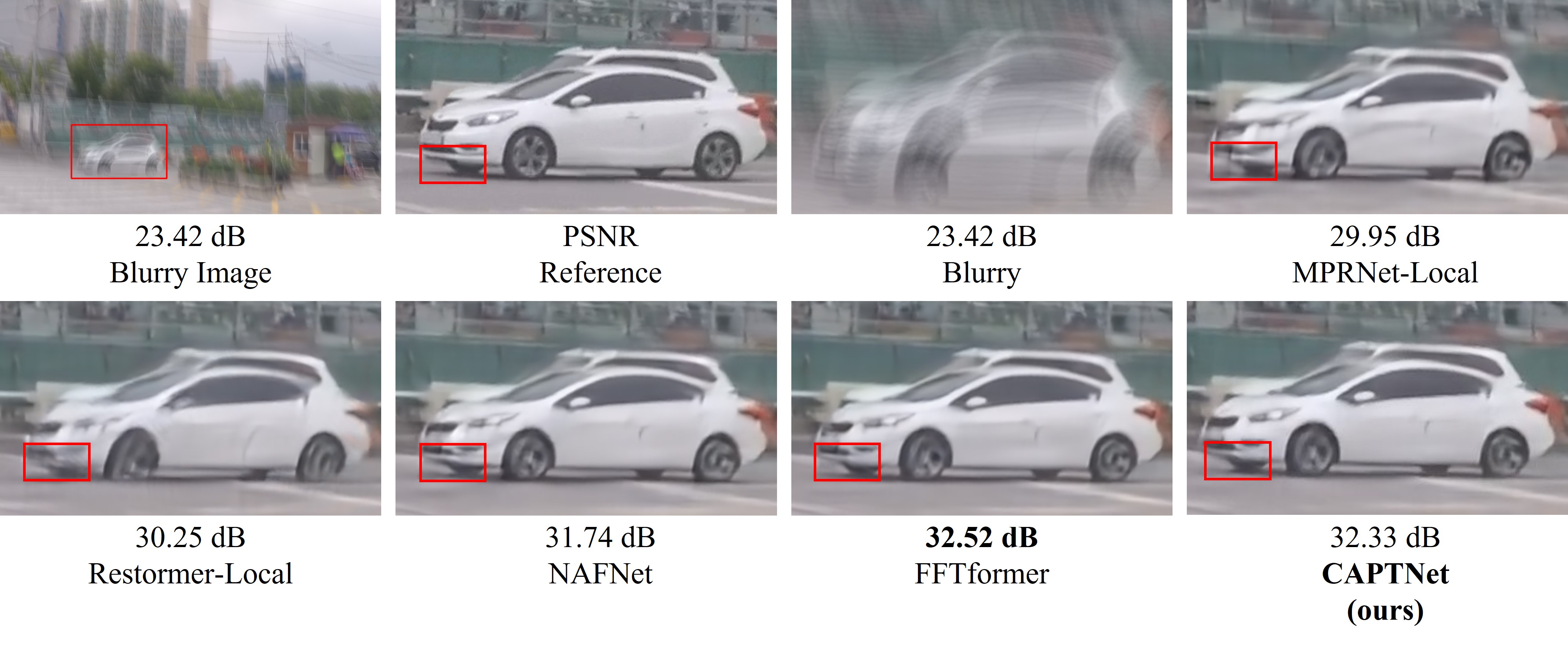}
	\caption{Visual under Task-Special restoration setting compare with state-of-the-art methods on GoPro dataset.}
	\label{fig:0blur_s}
\end{figure*}

In this section, we evaluate the performance of our CAPTNet under the task-special setting, i.e., separate models for different image restoration tasks.  The experimental results provide evidence that the proposed method effectively functions as a All-in-One model for restoring images with different types of degradation, while also serving as a model specifically designed for the single task. Table.~\ref{tb:01derain} shows the deraining result. Although our model, we did not achieve the best results when averaging four datasets. However, on all four datasets, our results are competitive with the best methods, $e.g.$, we obtain a performance gain of 1.15db over DRSformer~\cite{rainchen2023learning} and 1.58 db over Restormer~\cite{Zamir2021Restormer} (on Test1200). Fig.~\ref{fig:0rain_s}  shows visual comparisons on challenging images. We report the performance of evaluated image deblurring approaches on GoPro~\cite{Gopro} and HIDE~\cite{HIDE} datasets in Table.~\ref{tb:0deblurring}.  Compare to the previous best methods, we obtain the best on HIDE datasets, our CAPTNet yields 0.26 dB performance gain
over the previous best method FFTformer~\cite{blurkong2023efficient}, and 0.37 dB over the second best approach Restormer~\cite{Zamir2021Restormer}. It is worth noticing that our model is trained only on the GoPro datasets, thereby proving the strong generalization capability. While our approach is no match for the best approach FFTformer~\cite{blurkong2023efficient} on the GoPro dataset, it outperforms Restormer~\cite{Zamir2021Restormer}, NAFNet~\cite{chen2022simple}, HINet~\cite{Chen_2021_CVPRhinet} and Uformer~\cite{transWang_2022_CVPR_uform}.  We visualized the results of image deblurring in Fig.~\ref{fig:0blur_s}.

\begin{table*}[]
\centering
    \caption{Quantitative deraining results under Task-Special restoration setting with state-of-the-arts general image restoration.}
    \label{tb:01derain}
\begin{tabular}{ccccccccc||cc}
    \hline
    \multicolumn{1}{c}{} & \multicolumn{2}{c}{Test100~\cite{90Zhang2017ImageDU}}  & \multicolumn{2}{c}{Test1200~\cite{DIDMDN}} & \multicolumn{2}{c}{Rain100H~\cite{81Yang2016DeepJR}} & \multicolumn{2}{c||}{Rain100L~\cite{81Yang2016DeepJR}} & \multicolumn{2}{c}{Average} 
    \\
   Methods &PSNR $\uparrow$ &  SSIM $\uparrow$  & PSNR $\uparrow$ & SSIM $\uparrow$ &PSNR $\uparrow$ &SSIM $\uparrow$ & PSNR $\uparrow$&SSIM $\uparrow$ & PSNR & SSIM 
    \\
    \hline\hline
    DIDMDN~\cite{DIDMDN} & 22.56 & 0.818  & 29.65 & 0.901  & 17.35 &  0.524 &25.23 & 0.741 & 24.58 & 0.770
     \\
   MSPFN~\cite{MSPFN}  & 27.50 & 0.876 & 32.39 &  0.916   & 28.66 & 0.860  & 32.40 & 0.933 & 30.75  & 0.903
       \\
     MPRNet~\cite{Zamir2021MPRNet}  & 30.27 & 0.897 & 32.91 &  0.916   & 30.41 & 0.890  & 36.40 & 0.965 & 32.73  & 0.921
       \\
     SPAIR~\cite{SPAIR}  & 30.35 & 0.909& 33.04 &  0.922   & 30.95 & 0.892  & 36.93 & 0.969& 32.91 & \underline{0.926}
     \\
      HINet~\cite{Chen_2021_CVPRhinet}  & 30.29 & 0.906 & 33.05&  0.919   & 30.65 & 0.894  & 37.28 & 0.970 & 32.81 & 0.922
     \\
     Restormer~\cite{Zamir2021Restormer} &\textbf{32.00} & \textbf{0.923} & 33.19 & 0.926 & \underline{31.46} &\textbf{0.904} &38.99 &0.978 &33.91 & \textbf{0.933}
     \\
     Uformer~\cite{Zamir2021Restormer} &31.22 & 0.904 & 33.06 & 0.921 & 30.80 & 0.891 & 38.20 & 0.976 & 33.32 & 0.923 
     \\
     DRSformer~\cite{rainchen2023learning} & 31.71 & 0.911 & \underline{33.62} & \underline{0.929} & \textbf{32.03} & \underline{0.903} &\textbf{41.01} & \textbf{0.989} & \textbf{34.59} & \textbf{0.933}
     \\
     \hline
      \textbf{CAPTNet(ours)}  & \underline{31.77} & \underline{0.912} & \textbf{34.77}& \textbf{0.937}&31.37 & 0.897 &\underline{39.22} & \underline{0.981}   & \underline{34.28}    & \underline{0.926}
      \\
    \hline
\end{tabular}
\end{table*}

\begin{table}
\centering
\caption{Quantitative deblurring results under Task-Special restoration setting with state-of-the-arts general image restoration.}
\label{tb:0deblurring}
\begin{tabular}{ccccc}
    \hline
    \multicolumn{1}{c}{} & \multicolumn{2}{c}{GoPro~\cite{Gopro}}  & \multicolumn{2}{c}{HIDE~\cite{HIDE}} 
    \\
   Methods & PSNR $\uparrow$ & SSIM $\uparrow$ & PSNR $\uparrow$ & SSIM $\uparrow$   
    \\
    \hline\hline
    SPAIR~\cite{SPAIR} & 32.06 & 0.953 & 30.29 & 0.931 
    \\
    MIMO-UNet++~\cite{2021Rethinking} & 32.45 & 0.957 & 29.99 & 0.930 
    \\
    MPRNet~\cite{Zamir2021MPRNet} & 32.66 & 0.959 & 30.96 & 0.939 
    \\
    MPRNet-local~\cite{Zamir2021MPRNet} & 33.31 & 0.964 &31.19 &0.945 
    \\
    Uformer~\cite{transWang_2022_CVPR_uform} &32.97 & \underline{0.967} &30.83 &\textbf{0.952} 
     \\
    HINet~\cite{Chen_2021_CVPRhinet}&32.71&-&-&-
    \\
    HINet-local~\cite{Chen_2021_CVPRhinet}&33.08&0.962&-&-
    \\
     Restormer~\cite{Zamir2021Restormer} & 32.92 & 0.961 & 31.22 & 0.942 
    \\
    Restormer-local~\cite{Zamir2021Restormer} & 33.57 & 0.966& 31.49 & 0.945 
    \\
    NAFNet-local~\cite{chen2022simple}&33.69&\underline{0.967}&31.31&0.943
    \\
    FFTformer~\cite{blurkong2023efficient}& \textbf{34.21} & \textbf{0.969} & \underline{31.62} & 0.945 
    \\
    \hline
    \textbf{CAPTNet-32 (ours)}&33.07&0.962&31.08&0.941
    \\
    \textbf{CAPTNet-64 (ours)}&\underline{33.74}&\underline{0.967}&\textbf{31.86}&\underline{0.949}
    \\
    \hline
\end{tabular}
\end{table}

\subsubsection{\textbf{Unknown Task Results}}
Table.~\ref{tb:0unkonwn} evaluates the performance of each method on unknown tasks, i.e. the unknown noise level 100, without any fine-tuning. Our CAPTNet demonstrates the favorable
generalization ability compared to AirNet~\cite{airnetli2022all}, yielding 5.47 dB PSNR difference.

\begin{table}
\centering
\caption{Quantitative results of unknown tasks.}
\label{tb:0unkonwn}
\begin{tabular}{ccc}
    \hline
    \multicolumn{1}{c}{} &BSD68~\cite{BSDmartin2001database} & Urban100~\cite{urbanbsd7299156}
    \\
    \hline
   Methods & PSNR $\uparrow$ & PSNR $\uparrow$  
   \\
   Transweather~\cite{allvalanarasu2022transweather} & 13.04 & 13.01
   \\
   TAPE~\cite{all_liu2022tape} & 13.31 & 12.99
   \\
   AirNet~\cite{airnetli2022all} & 13.46 & 13.07
   \\
   \textbf{CAPTNet (ours)} &18.93 & 19.09
   \\
    \hline
\end{tabular}
\end{table}

\subsection{Abalation Studies}
Here we present ablation experiments to verify the effectiveness and scalability of our method. Evaluation is performed on the the combined dataset under All-in-One setting, and the results are shown in Table.~\ref{tb:ablex}. The metrics are reported on the average of all four datasets, from which we can make the following observations: \textbf{a)} Our model yields better performance after combine the Conv block and SPT block, which validates the effectiveness of our design. \textbf{b)} The MRAP is crucial for achieving overall performance improvement,  which leads to a gain of 0.55 dB. \textbf{c)} The proposed SGFN can selectively identify which frequency information needs to be retained,  resulting in a 0.07 dB improvement in our model. \textbf{d)} The FFM makes a gain of 0.02 dB for our model. \textbf{e)} When the MRAP, SGFN, and FFM are used as a single module in other networks, performance can be improved. When all three of them are used together, the lifting capacity of each module is higher.

In the hierarchical architecture of our CAPTNet, we analyzed the effect of the number of prompts and the position of the injection prompt on model performance. Table.~\ref{tb:ablprompt} shows that adding to decoder works better than encoder, and adding to level-4 works better than level-3. At the same time, it also shows that more is not always better, and the use of encoder and decoder at the same time can reduce model performance.

Table 5 evaluates the impact on CAPTNet performance by different combinations of degradation tasks, where the R, H, N, B,  denotes the derain, dehaze, denoise and deblur, respectively. With the inclusion of more tasks, it is evident that the performance remains stable or even demonstrates improvement, indicating its scalability when dealing with a range of degradation tasks.

\begin{table}
\caption{Ablation study on individual components of the
proposed CAPTNet.  Noted that "-" means that this module is not added, "\faTimes" means that the standard method is used.}
\label{tb:ablex}
    \centering
    \begin{tabular}{cccc|c}
         \hline
         Block Combination & MRAP & FFM & SGFN & PSNR
        \\
        \hline
        Conv  & - & - & - & 30.86
        \\
        Transformer & \faTimes & - &\faTimes & 31.96
        \\
        Conv + Transformer & \faTimes & -  & \faTimes  & 31.97
        \\
        Conv + Transformer & \faTimes & -  & \faCheck  & 32.04
        \\
        Conv + Transformer & \faCheck & -  & \faTimes  & 32.52
        \\
        Conv + Transformer & \faCheck & -  & \faCheck  & 32.61
        \\
        Conv + Transformer & \faTimes & \faCheck  & \faTimes  & 31.99
        \\
        Conv + Transformer & \faCheck & \faCheck  & \faTimes  & 32.55
        \\
        Conv + Transformer & \faTimes & \faCheck  & \faCheck  & 32.07
        \\
        Conv + Transformer & \faCheck & \faCheck  & \faCheck  & 32.65
         \\
         \hline
    \end{tabular}
\end{table}

\begin{table}
\caption{Ablation study on prompt number and injection position.}
\label{tb:ablprompt}
    \centering
    \begin{tabular}{cc|c}
         \hline
         Number & Position & PSNR
        \\
        \hline
        0  & - & 32.01
        \\
        1 &  level-3 encoder & 32.16
        \\
        1 & level-4 encoder & 32.22
        \\
        1 & level-3 decoder & 32.25
        \\
        1 & level-4 decoder & 32.34
        \\
        2 & level-3 and level-4 encoder & 32.47
        \\
        2 &  level-4 decoder and level-4 encoder & 32.42
        \\
        2 &  level-3 and level-4 decoder &32.65
        \\
        3 &  level-3 and level-4 decoder, level-4 encoder &32.59
        \\
         3 &  level-3 and level-4 decoder, level-3 encoder &32.56
         \\
         3 &  level-3 and level-4 encoder, level-4 decoder &32.45
        \\
         3 &  level-3 and level-4 encoder, level-3 decoder &32.38
         \\
         \hline
    \end{tabular}
\end{table}

\begin{table}
\caption{Ablation study on on different combinations of degradation tasks.}
\label{tb:abltype}
    \centering
    \begin{tabular}{cccc|c}
         \hline
         Tasks & GoPro & Rain100L & BSD68 & SOTS
        \\
        \hline
        B+R &32.78 &37.72 &21.20 & 19.82
        \\
         R+N+H & 27.81 & 37.53 & 30.25 & 29.29
         \\
         B+N+H & 32.65 & 33.41 & 30.13 & 29.22
         \\
         B+R+H & 32.70 & 37.52 & 21.45 & 29.25
         \\
         B+R+H+N & 32.71 & 37.86 & 30.75 & 29.28
         \\
         \hline
    \end{tabular}
\end{table}

\section{Discussion}
While the numerous experiments mentioned above have indeed confirmed the effectiveness of our method, it does not align perfectly with our initial motivation. In this section, we will systematically discussion the feasibility of our motivation. 

\begin{figure*}[htb] 
	\centering
	\includegraphics[width=0.9\textwidth]{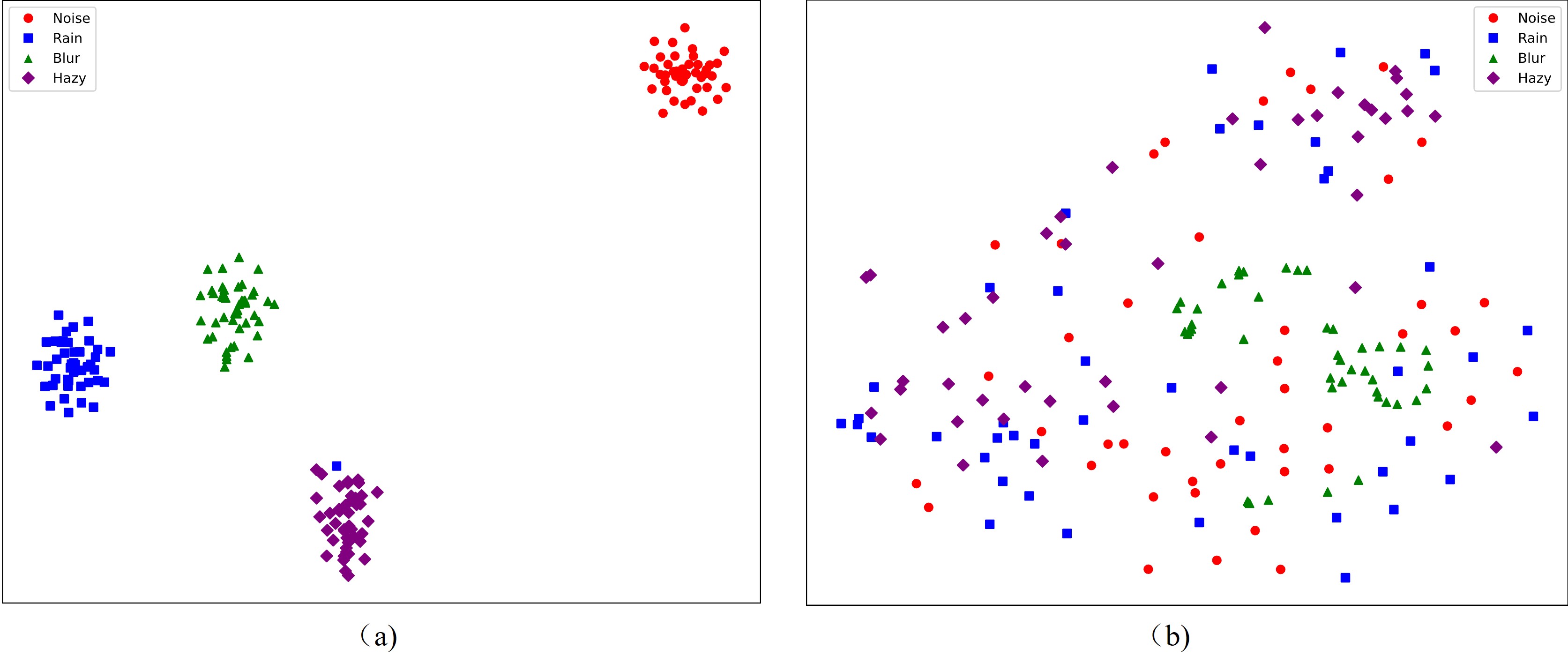}
	\caption{t-SNE visualization the output feature of our model for each degradation type of data. (a) shows the distribution of the output feture of level-3 SPT decoder. And (b) shows the distribution of final output feature of out model.}
	\label{fig:4capt_pro_fi}
\end{figure*}

In Sec.~\ref{introduction}, Fig.~\ref{fig:02tsne_dataset} shows that different types of degraded images have obvious distinctions in the feature space (the same degraded types are relatively clustered), while their corresponding high-quality images have no obvious differences in the feature space. This leads us to  formulate a hypothesis, with specific prompts, the model can efficiently handle diverse forms of image degradation, dynamically mapping the feature space of deteriorated images to that of high-quality images. Thus, we proposed  a data ingredient-oriented method  CAPTNet which injection the learnable prompts into SPT decoder. If our hypothesis is accurate, we can expect that the output features of images with different degradation types will exhibit noticeable distinctions after undergoing the decoder process with prompt-based information. We visualize the t-SNE statistics of the output of our SPT decoder and the final output of our model. As shown in Fig.~\ref{fig:4capt_pro_fi}(a), it has a more pronounced clustering phenomenon than the visualized result in Fig.~\ref{fig:4capt_middle}.  This substantiates the accuracy of our hypothesis, while the provided information in the prompt aids the model in discerning the various degradation types. In addition, our model can adaptive process images of various degradation types and effectively map them to a unified feature space (See Fig.~\ref{fig:4capt_pro_fi}(b)).

\section{Conclusion}
Most existing methods have been tailored for specific degradation types, limiting their applicability in real-world scenarios where diverse degradation types may be encountered. To overcome this limitation, we introduce a novel  data ingredient-oriented approach, leveraging prompt-based learning to empower a single model to efficiently address multiple image degradation tasks. The proposed  CNNs and Prompt Transformer network (CAPTNet) seamlessly combines CNN operations with Transformers to capture both local invariant properties and non-local information. Within the Transformer blocks, we implement several key designs, including multi-head rearranged attention with prompts and a simple-gate feed-forward network. These design  enable the model to selectively preserve information and intelligently guide the adaptive restoration of images affected by diverse degradations through the embedded learnable prompts. Furthermore, we introduce a feature fusion mechanism that explores multi-scale information to improve the aggregation of features. Extensive experiments validate the effectiveness and scalability of the proposed method.



\bibliographystyle{IEEEtran}
\bibliography{refbib}

\begin{thebibliography}{10}
\providecommand{\url}[1]{#1}
\csname url@samestyle\endcsname
\providecommand{\newblock}{\relax}
\providecommand{\bibinfo}[2]{#2}
\providecommand{\BIBentrySTDinterwordspacing}{\spaceskip=0pt\relax}
\providecommand{\BIBentryALTinterwordstretchfactor}{4}
\providecommand{\BIBentryALTinterwordspacing}{\spaceskip=\fontdimen2\font plus
\BIBentryALTinterwordstretchfactor\fontdimen3\font minus
  \fontdimen4\font\relax}
\providecommand{\BIBforeignlanguage}[2]{{%
\expandafter\ifx\csname l@#1\endcsname\relax
\typeout{** WARNING: IEEEtran.bst: No hyphenation pattern has been}%
\typeout{** loaded for the language `#1'. Using the pattern for}%
\typeout{** the default language instead.}%
\else
\language=\csname l@#1\endcsname
\fi
#2}}
\providecommand{\BIBdecl}{\relax}
\BIBdecl

\bibitem{noiseGuo2019Cbdnet}
S.~Guo, Z.~Yan, K.~Zhang, W.~Zuo, and L.~Zhang, ``Toward convolutional blind
  denoising of real photographs,'' \emph{2019 IEEE Conference on Computer
  Vision and Pattern Recognition (CVPR)}, 2019.

\bibitem{noiselee2022apbsn}
W.~Lee, S.~Son, and K.~M. Lee, ``Ap-bsn: Self-supervised denoising for
  real-world images via asymmetric pd and blind-spot network,'' in
  \emph{Proceedings of the IEEE/CVF Conference on Computer Vision and Pattern
  Recognition (CVPR)}, 2022.

\bibitem{noisezhang2023real}
Z.~Zhang, Y.~Jiang, W.~Shao, X.~Wang, P.~Luo, K.~Lin, and J.~Gu, ``Real-time
  controllable denoising for image and video,'' in \emph{Proceedings of the
  IEEE/CVF Conference on Computer Vision and Pattern Recognition}, 2023, pp.
  14\,028--14\,038.

\bibitem{blurkong2023efficient}
L.~Kong, J.~Dong, J.~Ge, M.~Li, and J.~Pan, ``Efficient frequency domain-based
  transformers for high-quality image deblurring,'' in \emph{Proceedings of the
  IEEE/CVF Conference on Computer Vision and Pattern Recognition}, 2023, pp.
  5886--5895.

\bibitem{blurpan2023cascaded}
J.~Pan, B.~Xu, H.~Bai, J.~Tang, and M.-H. Yang, ``Cascaded deep video
  deblurring using temporal sharpness prior and non-local spatial-temporal
  similarity,'' \emph{IEEE Transactions on Pattern Analysis and Machine
  Intelligence}, 2023.

\bibitem{blurpan2023deep}
J.~Pan, B.~Xu, J.~Dong, J.~Ge, and J.~Tang, ``Deep discriminative spatial and
  temporal network for efficient video deblurring,'' in \emph{Proceedings of
  the IEEE/CVF Conference on Computer Vision and Pattern Recognition}, 2023,
  pp. 22\,191--22\,200.

\bibitem{rain9798773}
J.~Xiao, X.~Fu, A.~Liu, F.~Wu, and Z.-J. Zha, ``Image de-raining transformer,''
  \emph{IEEE Transactions on Pattern Analysis and Machine Intelligence}, pp.
  1--18, 2022.

\bibitem{rainchen2023learning}
X.~Chen, H.~Li, M.~Li, and J.~Pan, ``Learning a sparse transformer network for
  effective image deraining,'' in \emph{Proceedings of the IEEE/CVF Conference
  on Computer Vision and Pattern Recognition}, 2023, pp. 5896--5905.

\bibitem{rainyang2023alternating}
D.~Yang, X.~He, and R.~Zhang, ``Alternating attention transformer for single
  image deraining,'' \emph{Digital Signal Processing}, p. 104144, 2023.

\bibitem{hazeli2017aod}
B.~Li, X.~Peng, Z.~Wang, J.~Xu, and D.~Feng, ``Aod-net: All-in-one dehazing
  network,'' in \emph{Proceedings of the IEEE international conference on
  computer vision}, 2017, pp. 4770--4778.

\bibitem{hazeqin2020ffa}
X.~Qin, Z.~Wang, Y.~Bai, X.~Xie, and H.~Jia, ``Ffa-net: Feature fusion
  attention network for single image dehazing,'' in \emph{Proceedings of the
  AAAI Conference on Artificial Intelligence}, vol.~34, no.~07, 2020, pp.
  11\,908--11\,915.

\bibitem{hazesong2022rethinking}
Y.~Song, Y.~Zhou, H.~Qian, and X.~Du, ``Rethinking performance gains in image
  dehazing networks,'' \emph{arXiv preprint arXiv:2209.11448}, 2022.

\bibitem{hazezheng2023curricular}
Y.~Zheng, J.~Zhan, S.~He, J.~Dong, and Y.~Du, ``Curricular contrastive
  regularization for physics-aware single image dehazing,'' in \emph{IEEE/CVF
  Conference on Computer Vision and Pattern Recognition}, 2023.

\bibitem{IPT}
H.~Chen, Y.~Wang, T.~Guo, C.~Xu, Y.~Deng, Z.~Liu, S.~Ma, C.~Xu, C.~Xu, and
  W.~Gao, ``Pre-trained image processing transformer,'' \emph{2021 IEEE/CVF
  Conference on Computer Vision and Pattern Recognition (CVPR)}, pp.
  12\,294--12\,305, 2020.

\bibitem{li2020all}
R.~Li, R.~T. Tan, and L.-F. Cheong, ``All in one bad weather removal using
  architectural search,'' in \emph{Proceedings of the IEEE/CVF conference on
  computer vision and pattern recognition}, 2020, pp. 3175--3185.

\bibitem{valanarasu2022transweather}
J.~M.~J. Valanarasu, R.~Yasarla, and V.~M. Patel, ``Transweather:
  Transformer-based restoration of images degraded by adverse weather
  conditions,'' in \emph{Proceedings of the IEEE/CVF Conference on Computer
  Vision and Pattern Recognition}, 2022, pp. 2353--2363.

\bibitem{liu2022tape}
L.~Liu, L.~Xie, X.~Zhang, S.~Yuan, X.~Chen, W.~Zhou, H.~Li, and Q.~Tian,
  ``Tape: Task-agnostic prior embedding for image restoration,'' in
  \emph{European Conference on Computer Vision}.\hskip 1em plus 0.5em minus
  0.4em\relax Springer, 2022, pp. 447--464.

\bibitem{airnetli2022all}
B.~Li, X.~Liu, P.~Hu, Z.~Wu, J.~Lv, and X.~Peng, ``All-in-one image restoration
  for unknown corruption,'' in \emph{Proceedings of the IEEE/CVF Conference on
  Computer Vision and Pattern Recognition}, 2022, pp. 17\,452--17\,462.

\bibitem{IDRzhang2023ingredient}
J.~Zhang, J.~Huang, M.~Yao, Z.~Yang, H.~Yu, M.~Zhou, and F.~Zhao,
  ``Ingredient-oriented multi-degradation learning for image restoration,'' in
  \emph{Proceedings of the IEEE/CVF Conference on Computer Vision and Pattern
  Recognition}, 2023, pp. 5825--5835.

\bibitem{chen2022simple}
L.~Chen, X.~Chu, X.~Zhang, and J.~Sun, ``Simple baselines for image
  restoration,'' \emph{arXiv preprint arXiv:2204.04676}, 2022.

\bibitem{Gopro}
S.~Nah, T.~H. Kim, and K.~M. Lee, ``Deep multi-scale convolutional neural
  network for dynamic scene deblurring,'' \emph{2017 IEEE Conference on
  Computer Vision and Pattern Recognition (CVPR)}, pp. 257--265, 2016.

\bibitem{81Yang2016DeepJR}
W.~Yang, R.~T. Tan, J.~Feng, J.~Liu, Z.~Guo, and S.~Yan, ``Deep joint rain
  detection and removal from a single image,'' \emph{CVPR}, 2017.

\bibitem{BSDmartin2001database}
D.~Martin, C.~Fowlkes, D.~Tal, and J.~Malik, ``A database of human segmented
  natural images and its application to evaluating segmentation algorithms and
  measuring ecological statistics,'' in \emph{Proceedings Eighth IEEE
  International Conference on Computer Vision. ICCV 2001}, vol.~2.\hskip 1em
  plus 0.5em minus 0.4em\relax IEEE, 2001, pp. 416--423.

\bibitem{RESIDEli2018benchmarking}
B.~Li, W.~Ren, D.~Fu, D.~Tao, D.~Feng, W.~Zeng, and Z.~Wang, ``Benchmarking
  single-image dehazing and beyond,'' \emph{IEEE Transactions on Image
  Processing}, vol.~28, no.~1, pp. 492--505, 2018.

\bibitem{1992Nonlinear}
L.~I. Rudin, S.~Osher, and E.~Fatemi, ``Nonlinear total variation based noise
  removal algorithms,'' \emph{Physica D Nonlinear Phenomena}, 1992.

\bibitem{1997Prior}
C.~Z. Song and D.~Mumford, ``Prior learning and gibbs reaction-diffusion,''
  \emph{TPAMI}, vol.~19, no.~11, pp. 1236--1250, 1997.

\bibitem{2005Fields}
S.~Roth and M.~J. Black, ``Fields of experts: A framework for learning image
  priors,'' in \emph{CVPR}, 2005.

\bibitem{2011Image}
W.~Dong, L.~Zhang, G.~Shi, and X.~Wu, ``Image deblurring and super-resolution
  by adaptive sparse domain selection and adaptive regularization,''
  \emph{TIP}, vol.~20, no.~7, pp. 1838--1857, 2011.

\bibitem{2011Single}
K.~He, J.~Sun, and X.~Tang, ``Single image haze removal using dark channel
  prior,'' \emph{TPAMI}, 2011.

\bibitem{cnnspairpurohit2021spatially}
K.~Purohit, M.~Suin, A.~Rajagopalan, and V.~N. Boddeti, ``Spatially-adaptive
  image restoration using distortion-guided networks,'' in \emph{Proceedings of
  the IEEE/CVF International Conference on Computer Vision}, 2021, pp.
  2309--2319.

\bibitem{Zamir2022MIRNetv2}
S.~W. Zamir, A.~Arora, S.~Khan, M.~Hayat, F.~S. Khan, M.-H. Yang, and L.~Shao,
  ``Learning enriched features for fast image restoration and enhancement,''
  \emph{TPAMI}, 2022.

\bibitem{Zamir2021MPRNet}
------, ``Multi-stage progressive image restoration,'' in \emph{CVPR}, 2021.

\bibitem{cnnmou2022deep}
C.~Mou, Q.~Wang, and J.~Zhang, ``Deep generalized unfolding networks for image
  restoration,'' in \emph{Proceedings of the IEEE/CVF Conference on Computer
  Vision and Pattern Recognition}, 2022, pp. 17\,399--17\,410.

\bibitem{cnnpan2022dual}
J.~Pan, D.~Sun, J.~Zhang, J.~Tang, J.~Yang, Y.-W. Tai, and M.-H. Yang, ``Dual
  convolutional neural networks for low-level vision,'' \emph{International
  Journal of Computer Vision}, vol. 130, no.~6, pp. 1440--1458, 2022.

\bibitem{transchen2021preipt}
H.~Chen, Y.~Wang, T.~Guo, C.~Xu, Y.~Deng, Z.~Liu, S.~Ma, C.~Xu, C.~Xu, and
  W.~Gao, ``Pre-trained image processing transformer,'' in \emph{Proceedings of
  the IEEE/CVF conference on computer vision and pattern recognition}, 2021,
  pp. 12\,299--12\,310.

\bibitem{transli2020learning}
X.~Li, X.~Jin, J.~Lin, S.~Liu, Y.~Wu, T.~Yu, W.~Zhou, and Z.~Chen, ``Learning
  disentangled feature representation for hybrid-distorted image restoration,''
  in \emph{Computer Vision--ECCV 2020: 16th European Conference, Glasgow, UK,
  August 23--28, 2020, Proceedings, Part XXIX 16}.\hskip 1em plus 0.5em minus
  0.4em\relax Springer, 2020, pp. 313--329.

\bibitem{transliang2021swinir}
J.~Liang, J.~Cao, G.~Sun, K.~Zhang, L.~Van~Gool, and R.~Timofte, ``Swinir:
  Image restoration using swin transformer,'' in \emph{Proceedings of the
  IEEE/CVF international conference on computer vision}, 2021, pp. 1833--1844.

\bibitem{Zamir2021Restormer}
S.~W. Zamir, A.~Arora, S.~Khan, M.~Hayat, F.~S. Khan, and M.-H. Yang,
  ``Restormer: Efficient transformer for high-resolution image restoration,''
  in \emph{CVPR}, 2022.

\bibitem{transWang_2022_CVPR_uform}
Z.~Wang, X.~Cun, J.~Bao, W.~Zhou, J.~Liu, and H.~Li, ``Uformer: A general
  u-shaped transformer for image restoration,'' in \emph{Proceedings of the
  IEEE/CVF Conference on Computer Vision and Pattern Recognition (CVPR)}, June
  2022, pp. 17\,683--17\,693.

\bibitem{all_multie_li2020all}
R.~Li, R.~T. Tan, and L.-F. Cheong, ``All in one bad weather removal using
  architectural search,'' in \emph{Proceedings of the IEEE/CVF conference on
  computer vision and pattern recognition}, 2020, pp. 3175--3185.

\bibitem{allvalanarasu2022transweather}
J.~M.~J. Valanarasu, R.~Yasarla, and V.~M. Patel, ``Transweather:
  Transformer-based restoration of images degraded by adverse weather
  conditions,'' in \emph{Proceedings of the IEEE/CVF Conference on Computer
  Vision and Pattern Recognition}, 2022, pp. 2353--2363.

\bibitem{all_conli2022all}
B.~Li, X.~Liu, P.~Hu, Z.~Wu, J.~Lv, and X.~Peng, ``All-in-one image restoration
  for unknown corruption,'' in \emph{Proceedings of the IEEE/CVF Conference on
  Computer Vision and Pattern Recognition}, 2022, pp. 17\,452--17\,462.

\bibitem{all_liu2022tape}
L.~Liu, L.~Xie, X.~Zhang, S.~Yuan, X.~Chen, W.~Zhou, H.~Li, and Q.~Tian,
  ``Tape: Task-agnostic prior embedding for image restoration,'' in
  \emph{European Conference on Computer Vision}.\hskip 1em plus 0.5em minus
  0.4em\relax Springer, 2022, pp. 447--464.

\bibitem{gptfloridi2020gpt}
L.~Floridi and M.~Chiriatti, ``Gpt-3: Its nature, scope, limits, and
  consequences,'' \emph{Minds and Machines}, vol.~30, pp. 681--694, 2020.

\bibitem{vptjia2022visual}
M.~Jia, L.~Tang, B.-C. Chen, C.~Cardie, S.~Belongie, B.~Hariharan, and S.-N.
  Lim, ``Visual prompt tuning,'' in \emph{European Conference on Computer
  Vision}.\hskip 1em plus 0.5em minus 0.4em\relax Springer, 2022, pp. 709--727.

\bibitem{pronie2022pro}
X.~Nie, B.~Ni, J.~Chang, G.~Meng, C.~Huo, Z.~Zhang, S.~Xiang, Q.~Tian, and
  C.~Pan, ``Pro-tuning: Unified prompt tuning for vision tasks,'' \emph{arXiv
  preprint arXiv:2207.14381}, 2022.

\bibitem{lionwang2023lion}
H.~Wang, J.~Chang, X.~Luo, J.~Sun, Z.~Lin, and Q.~Tian, ``Lion: Implicit vision
  prompt tuning,'' \emph{arXiv preprint arXiv:2303.09992}, 2023.

\bibitem{vaswani2017attention}
A.~Vaswani, N.~Shazeer, N.~Parmar, J.~Uszkoreit, L.~Jones, A.~N. Gomez,
  {\L}.~Kaiser, and I.~Polosukhin, ``Attention is all you need,''
  \emph{Advances in neural information processing systems}, vol.~30, 2017.

\bibitem{vitdosovitskiy2020image}
A.~Dosovitskiy, L.~Beyer, A.~Kolesnikov, D.~Weissenborn, X.~Zhai,
  T.~Unterthiner, M.~Dehghani, M.~Minderer, G.~Heigold, S.~Gelly \emph{et~al.},
  ``An image is worth 16x16 words: Transformers for image recognition at
  scale,'' \emph{ICLR}, 2021.

\bibitem{238099669}
X.~Fu, J.~Huang, D.~Zeng, Y.~Huang, X.~Ding, and J.~Paisley, ``Removing rain
  from single images via a deep detail network,'' in \emph{CVPR}, 2017.

\bibitem{487780668}
Y.~Li, R.~T. Tan, X.~Guo, J.~Lu, and M.~S. Brown, ``Rain streak removal using
  layer priors,'' in \emph{CVPR}, 2016.

\bibitem{90Zhang2017ImageDU}
H.~Zhang, V.~A. Sindagi, and V.~M. Patel, ``Image de-raining using a
  conditional generative adversarial network,'' \emph{IEEE Transactions on
  Circuits and Systems for Video Technology}, vol.~30, pp. 3943--3956, 2017.

\bibitem{DIDMDN}
H.~Zhang and V.~M. Patel, ``Density-aware single image de-raining using a
  multi-stream dense network,'' \emph{2018 IEEE/CVF Conference on Computer
  Vision and Pattern Recognition}, pp. 695--704, 2018.

\bibitem{HIDE}
Z.~Shen, W.~Wang, X.~Lu, J.~Shen, H.~Ling, T.~Xu, and L.~Shao, ``Human-aware
  motion deblurring,'' \emph{2019 IEEE/CVF International Conference on Computer
  Vision (ICCV)}, pp. 5571--5580, 2019.

\bibitem{wedma2016waterloo}
K.~Ma, Z.~Duanmu, Q.~Wu, Z.~Wang, H.~Yong, H.~Li, and L.~Zhang, ``Waterloo
  exploration database: New challenges for image quality assessment models,''
  \emph{IEEE Transactions on Image Processing}, vol.~26, no.~2, pp. 1004--1016,
  2016.

\bibitem{2014Adam}
D.~Kingma and J.~Ba, ``Adam: A method for stochastic optimization,''
  \emph{Computer Science}, 2014.

\bibitem{2016SGDR}
I.~Loshchilov and F.~Hutter, ``Sgdr: Stochastic gradient descent with warm
  restarts,'' 2016.

\bibitem{Chu2021ImprovingIR}
X.~Chu, L.~Chen, C.~Chen, and X.~Lu, ``Improving image restoration by
  revisiting global information aggregation,'' in \emph{ECCV}, 2021.

\bibitem{Chen_2021_CVPRhinet}
L.~Chen, X.~Lu, J.~Zhang, X.~Chu, and C.~Chen, ``Hinet: Half instance
  normalization network for image restoration,'' in \emph{Proceedings of the
  IEEE/CVF Conference on Computer Vision and Pattern Recognition (CVPR)
  Workshops}, June 2021, pp. 182--192.

\bibitem{fan2019general}
Q.~Fan, D.~Chen, L.~Yuan, G.~Hua, N.~Yu, and B.~Chen, ``A general decoupled
  learning framework for parameterized image operators,'' \emph{IEEE
  transactions on pattern analysis and machine intelligence}, vol.~43, no.~1,
  pp. 33--47, 2019.

\bibitem{MSPFN}
K.~Jiang, Z.~Wang, P.~Yi, C.~Chen, B.~Huang, Y.~Luo, J.~Ma, and J.~Jiang,
  ``Multi-scale progressive fusion network for single image deraining,''
  \emph{CVPR}, 2020.

\bibitem{SPAIR}
K.~Purohit, M.~Suin, A.~N. Rajagopalan, and V.~N. Boddeti, ``Spatially-adaptive
  image restoration using distortion-guided networks,'' \emph{CoRR}, vol.
  abs/2108.08617, 2021.

\bibitem{2021Rethinking}
S.~J. Cho, S.~W. Ji, J.~P. Hong, S.~W. Jung, and S.~J. Ko, ``Rethinking
  coarse-to-fine approach in single image deblurring,'' in \emph{ICCV}, 2021.

\bibitem{urbanbsd7299156}
J.-B. Huang, A.~Singh, and N.~Ahuja, ``Single image super-resolution from
  transformed self-exemplars,'' in \emph{2015 IEEE Conference on Computer
  Vision and Pattern Recognition (CVPR)}, 2015, pp. 5197--5206.

\end{thebibliography}

\vfill

\end{document}